\pdfoutput=1
\documentclass[journal]{IEEEtran}
\IEEEoverridecommandlockouts
\usepackage{cite}
\usepackage{amsmath,amssymb,amsfonts}
\usepackage{algorithmic}
\usepackage{graphicx}
\usepackage{textcomp}
\usepackage{xcolor, colortbl}
\usepackage{times}
\usepackage{soul}
\usepackage{url}
\usepackage{adjustbox}
\usepackage{tabularx}
\usepackage{xspace}
\usepackage[normalem]{ulem}

\usepackage[hidelinks]{hyperref}
\usepackage[utf8]{inputenc}
\usepackage[small]{caption}
\usepackage{booktabs}
\usepackage{algorithm}
\usepackage{algorithmic}
\usepackage{multirow}
\usepackage{array}
\usepackage{subfigure}
\usepackage{bm}
\newtheorem{theorem}{Theorem}

\newcommand{\ourModelName}{CAGNNs\xspace}
\newcommand{\Q}[1]{\textcolor{black}{{#1}}}

\def\BibTeX{{\rm B\kern-.05em{\sc i\kern-.025em b}\kern-.08em
    T\kern-.1667em\lower.7ex\hbox{E}\kern-.125emX}}
\begin{document}

\title{Exploiting Neighbor Effect: Conv-Agnostic GNN Framework for Graphs with Heterophily\\

}

\author{
\thanks{This work is supported by the Shanghai Municipal Science and Technology Major Project (No. 2018SHZDZX01), National Natural Science Foundation of China (No. 62176059), and ZJLab, Shanghai Center for Brain Science and Brain-Inspired Technology. (Corresponding author: Jian Pu).}
\thanks{Jie Chen, Shouzhen Chen, and Junping Zhang are with Shanghai Key Lab of Intelligent Information Processing and the School of Computer Science, Fudan University, Shanghai 200433, China (email: \{chenj19, chensz19, jpzhang\}@fudan.edu.cn).}
\thanks{Junbin Gao is with Discipline of Business Analytics, the University of Sydney Business School, the University of Sydney, Camperdown,
NSW 2006, Australia (e-mail:
junbin.gao@sydney.edu.au)}
\thanks{Zengfeng Huang is with the School of Data Science, Fudan University,
Shanghai, 200433, China (e-mail:
huangzf@fudan.edu.cn)}
\thanks{Jian Pu is with the Institute of Science and Technology for Brain-Inspired
Intelligence, Fudan University,
Shanghai, 200433, China (e-mail:
jianpu@fudan.edu.cn)}

\normalsize{
Jie Chen,
Shouzhen Chen,
Junbin Gao,
Zengfeng Huang,
Junping Zhang,~\IEEEmembership{Member, IEEE}, 
Jian Pu
}
}

\markboth{IEEE TRANSACTIONS ON NEURAL NETWORKS AND LEARNING SYSTEMS, VOL. XX, NO. XX, XXX 2023}%
{Shell \MakeLowercase{\textit{et al.}}: Bare Demo of IEEEtran.cls for Journals}

\maketitle
\begin{abstract}
Due to the homophily assumption in graph convolution networks (GNNs), a common consensus in the graph node classification task is that GNNs perform well on homophilic graphs but may fail on heterophilic graphs with many inter-class edges. However, the previous inter-class edges perspective and related homo-ratio metrics cannot well explain the GNNs performance under some heterophilic datasets, which implies that not all the inter-class edges are harmful to GNNs. In this work, we propose a new metric based on von Neumann entropy to re-examine the heterophily problem of GNNs and investigate the feature aggregation of inter-class edges from an entire neighbor identifiable perspective. Moreover, we propose a simple yet effective Conv-Agnostic GNN framework (CAGNNs) to enhance the performance of most GNNs on heterophily datasets by learning the neighbor effect for each node. Specifically, we first decouple the feature of each node into the discriminative feature for downstream tasks and the aggregation feature for graph convolution. Then, we propose a shared mixer module to adaptively evaluate the neighbor effect of each node to incorporate the neighbor information. The proposed framework can be regarded as a plug-in component and is compatible with most GNNs. The experimental results over nine well-known benchmark datasets indicate that our framework can significantly improve performance, especially for the heterophily graphs. The average performance gain is 9.81\%, 25.81\%, and 20.61\% compared with GIN, GAT, and GCN, respectively. Extensive ablation studies and robustness analysis further verify the effectiveness, robustness, and interpretability of our framework. Code is available at \url{https://github.com/JC-202/CAGNN}.

\end{abstract}

\begin{IEEEkeywords}
Graph neural networks, Node classification, Representation learning, Heterophily, Homophily
\end{IEEEkeywords}

\section{{Introduction}}
Recently, graph neural networks (GNNs) have demonstrated a remarkable ability for graph node classification tasks. 
Most GNNs learn node representation by recursively aggregating neighbor information~\cite{tnnls-compre,Hamilton2017InductiveRL,bacciu2020gentle} and follow the well-known homophily assumption, i.e., nodes connected by edges tend to be of the same class. 
The aggregation process over intra-class edges would smooth representations of nodes from the same class, thus benefiting the classification task and achieving promising results on homophilic graphs~\cite{li2018deeper,nt2019revisiting,GCN}.
However, the opposite heterophily property is also observed in a wide range of real-world graphs, including dating~\cite{zhu2020beyond}, molecular~\cite{guo2022deep,ji2021graph,bacciu2022explaining}, and transaction networks~\cite{liu2021item, zheng2022graph}.
Unfortunately, the applications of GNNs in heterophilic graphs whose connected nodes often have different labels are usually problematic~\cite{yan2021two,bo2021beyond,zhu2021graph,zheng2022graph}. 

Most literature agrees that massive inter-class edges in heterophilic graphs are harmful for aggregation since they may {cause the graph convolution to over-smooth the representations of nodes belonging to different classes}~\cite{zhu2021graph,zheng2022graph,chen2022memory}.
Furthermore, they measure the proportion of inter-class edges in a graph as homo-ratio metrics to evaluate the "strength" of heterophily~\cite{pei2019geom,zhu2020beyond}, which can be used to guide the application of GNNs. However, such metrics from edge perspectives fail to explain the completely different node classification performance under some heterophilic graphs with a similar homo-ratio~\cite{pei2019geom, zhu2020beyond, luan2021heterophily}. 
Therefore, it compels us to re-examine the heterophily problem of GNNs and answer the following two questions: (1) \textit{Is inter-class neighborhood aggregation truly harmful or unnecessary?} (2) \textit{How to improve GNNs' performance on heterophilic graphs?}

\begin{figure}[!t]
\begin{minipage}{1.0\linewidth}
  \centering
  \centerline{\includegraphics[width=8.5cm]{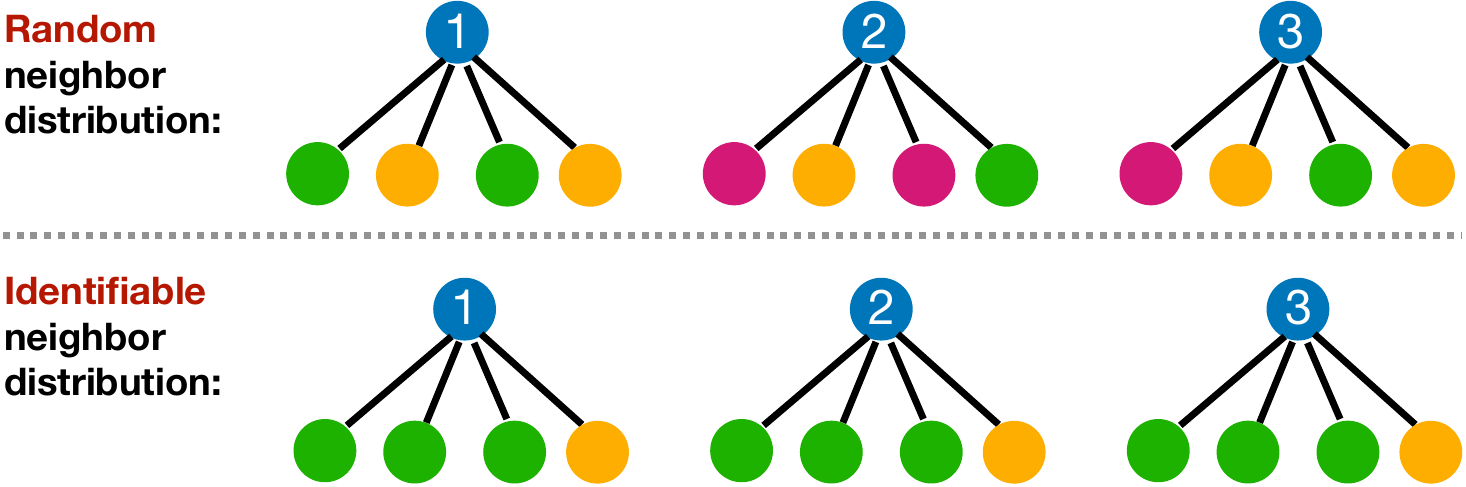}}
\end{minipage}
\caption{The \Q{identifiability of neighbors}. Nodes with the same colors share the same class labels. The above neighbor distribution for the blue class is random, and the bottom is not random. Although many inter-class edges exist, the identifiable patterns from the non-random distribution of dissimilar neighbors may also help graph convolution classify the center node's label.}
\label{fig:intro}
\end{figure}

As shown in Fig.~\ref{fig:intro}, we first note that the inter-class edges may be beneficial to improve the node classification task if their neighbor distribution is identifiable instead of random.
The critical point is to treat the neighbor as an entirety and measure the identifiability of each class neighbor distribution rather than simply considering the proportion of inter-class edges. 
A class with a random neighbor distribution is considered to have lower identifiability, 
{e.g., all nodes are randomly connected to others without considering classes of nodes.} 
Whereas a specific neighbor-connected pattern is considered to have higher identifiability.
An extreme example is a bipartite graph. Although the graph is highly heterophilic 
(each edge connects two nodes of the opposite class), node features are still distinguishable after a simple mean aggregation operator since their neighbor's distribution is identifiable~\cite{chien2020adaptive,ma2022is}. 

In this work, we first present a novel metric to quantify the identifiability of the neighbor's distribution. Compared with the previous inter-class edges metrics, ours can better indicate the graph-level neighbor effect for classifications and understand the heterophily problem. Then, motivated by this metric, we propose a simple yet effective Conv-Agnostic GNN Framework (\ourModelName) to improve GNNs' performance on heterophilic graphs by learning the node-level neighbor effect. We summarize the main contribution as follows.

\textit{(1) We present a new perspective of heterophily with neighbor identifiability and quantify it with the metric inspired by the von Neumann entropy.}
Specifically, we extract the label distribution matrix of neighbors at the class level and extend the von 
Neumann entropy~\cite{bengtsson2017geometry} as a new metric to 
measure the identifiability of neighbors for the graph. 
Unlike the edge perspective, this metric demonstrates the importance of the neighbor effect for classifications in understanding heterophily problems. It can be used to explain the performance differences of GNNs among various datasets with the same homo-ratio. Furthermore, it motivates us to consider the node-level neighbor effect to improve classical GNNs. 

\textit{(2) We propose \ourModelName as a general framework to improve classical GNNs by learning the \Q{neighbor effect} for each node.}
To consider the neighbor effect at the node level, we first decouple the features of a node into two parts: discriminative features used for downstream tasks and aggregated features from its neighbors via graph convolution. Then, we introduce a mixer module to fuse these two features into new discriminative features for classification via learning the node-level neighbor effect. Our framework can improve the performance of most GNNs under heterophily without modifying the graph convolution kernel due to the generic node-level neighbor effect learning mechanism. Thus, unlike other heterophilic-oriented GNNs that need to modify the convolution kernel~\cite{bo2021beyond,pei2019geom,zhu2020beyond}, our framework is conv-agnostic and more general.

\textit{(3) We conduct extensive experiments on nine well-known benchmark datasets to verify the effectiveness, interpretability, and robustness of \ourModelName.} We show that a simple shared-parameters mixer module can substantially enhance the performance of most GNNs on heterophilic graphs while maintaining the performance on homophilic graphs. The average performance gain is 9.81\%, 25.81\%, and 20.61\% compared with GIN, GAT, and GCN, respectively. Moreover, our framework shows good interpretability for determining whether neighbor information is helpful for node classification tasks, and the framework is robust to over-smoothing and noisy edges.

The paper is organized as follows.
In Section~\ref{Sec:2}, we survey the related works.
Section~\ref{Sec:3} briefly introduces the notations and background.
In Section~\ref{section:Metric}, 
we present the proposed novel metric for measuring neighbor identifiability for graphs with heterophily from a neighbor perspective.
In Section~\ref{Sec:5}, we describe the proposed CAGNN framework and implementation in detail. 
Evaluation results on nine benchmark datasets and ablation studies are presented in Section~\ref{Sec:6} to verify the effectiveness, interpretability, and robustness of the proposed framework. The final section concludes the paper.
\section{Related Works}\label{Sec:2}
\subsection{Graph Neural Networks}
GNN models can be roughly categorized into spectral and spatial methods. 
Early on, Bruna et al. \cite{bruna2013spectral} first propose a spectral graph-based extension of convolutional networks to graphs. In a follow-up work, ChebyNets \cite{defferrard2016convolutional} define graph convolutions using Chebyshev polynomials to remove the computationally expensive Laplacian eigendecomposition.
GCNs \cite{GCN} further simplify graph convolutions by stacking layers of first-order Chebyshev polynomial filters with a redefined propagation matrix. 
Also, the GCN bridges the spectral and spatial domain gap since it can also be regarded as a mean aggregator to aggregate neighbor information to each node. Furthermore, in the spatial domain, Graph Attention Network (GAT) applies the attention mechanism to learn edge weights to improve the aggregation step. Xu et al. \cite{xu2018powerful} study the expressiveness of graph neural networks and introduces Graph Isomorphism Network (GIN), which is proven to be as powerful as the Weisfeiler-Lehman test. 
There are many other graph neural models~\cite{klicpera2018predict,Hamilton2017InductiveRL,errica2019fair}; we refer to \cite{battaglia2018relational, tnnls-compre, bacciu2020gentle} for a more comprehensive review.

To investigate why and when graph neural networks work well for node classification, some researchers aim to understand the behavior of GNNs. Ni and Machara \cite{nt2019revisiting} indicate that graph neural networks only perform low-pass filtering on feature vectors and do not have the non-linear manifold learning property from a signal processing perspective. Li et al. \cite{li2018deeper} point out that the GCN model's graph convolution is actually a special form of Laplacian smoothing, which is consistent with the homophily assumption. However, it also brings potential concerns about making the features of connected nodes from different labels indistinguishable. On the other hand, Ma et al. \cite{ma2022is} theoretically reveal that homophily is not a necessary assumption for the GCN model.
Moreover, for attention-based GNNs, Wang et al. \cite{wang2019improving} find that stacking multiple attention layers causes excessive smoothing of node features due to information exchange over inter-class edges. 
\cite{hou2020measuring,chen2021graph} summarize the current popular message-passing scheme in GNNs and argue that the message between intra-class edges would help nodes receive information gain. In contrast, the inter-class edges may introduce negative disturbance. 

\subsection{GNNs for Heterophily}
Recently, heterophilic graph learning has become an upward-trending research topic, and various specific structured GNNs have been proposed. Most argue that message passing during inter-class edges is harmful to the node classification task and try to avoid harmfulness. Current kinds of literature can be divided into three lines: 

(1) Some works deal with heterophily from the spectral domain. FAGCN~\cite{bo2021beyond} divides the message from each edge into low-frequency and high-frequency signals and shows that both the low- and high-frequency signals are necessary for heterophilic graph learning. In addition, some works aim at extracting high-order approximation with graph spectral filters. GPRGNN~\cite{chien2020adaptive} modifies the convolution to the generalized page rank and learns an arbitrary K-order polynomial graph filter. GCNII~\cite{chen2020simple} proposes the initial residual and identity mapping for vanilla GCN and theoretically proves that it can express a K-order polynomial filter with arbitrary coefficients.
BernNet~\cite{he2021bernnet} learns arbitrary graph spectral filters via Bernstein approximation to oversimplified or ill-posed filters. 
{ACM-GCN~\cite{luan2021heterophily} modifies graph convolution by explicitly dividing it into low-pass, high-pass, and identity filters in each layer and adaptively fusing them for each node. However, these spectral-based methods need to specifically design and modify the graph convolution, which is not trivial to generalize to broader spatial GNNs like GAT and GIN.}

(2) Some works reorganize the graph structure to obtain a more homophilic signal.
Geom-GCN~\cite{pei2019geom} utilizes geometric aggregation to capture structural similarity in the latent space and long-range dependencies.
NLGNN~\cite{liu2021non} leverages attention-guide sorting to generate a re-connected graph and conducts non-local aggregation. 
SLAPS~\cite{fatemi2021slaps} combines the self-supervised technique to infer a homophily latent structure.
GDAMN~\cite{chen2021graph} proposes the decoupling attention mechanism on both features and labels to increase the intra-class and reduce inter-class edge weights. 

(3) Some works aim to capture high-order neighbor information, which was proven to be homophily dominant~\cite{bo2021beyond}.
%some work are aim to capture high order neighborhood
MixHop~\cite{abu2019mixhop} repeatedly mixes representations of multi-hop neighbors to achieve higher-order message passing.
JK-Nets~\cite{xu2018representation} jumps the intermediate representations to the last layer for better structure-aware
representation.
H2GCN~\cite{zhu2020beyond} proposes three designs with separated ego and neighbors, hider-order neighbors, and a combination of intermediate representations to combine the message from neighbors.

Unlike these specific GNN architectures to avoid the harmfulness of inter-class edges; instead, we consider the inter-class edges from an identifiable neighbor distribution perspective. Furthermore, we propose a simple and general Conv-Agnostic framework. This framework can be regarded as a plug-in component and is compatible with most GNNs to improve their performance on heterophilic graphs.

\section{Preliminary}\label{Sec:3}
\subsection{Problem Setup}
Consider an undirected graph $\cal G = (V, E)$ with adjacency matrix $\mathbf{A}\in \mathbb{R}^{N\times N}$ and the diagonal degree matrix $\mathbf{D}$ of $\mathbf{A}$, 
where $\cal V$ and $\cal E$ are the sets of nodes and edges, respectively.
For each node $v_i\in {\cal V}$, we denote $\mathcal{N}({v_i})=\{j:(i,j)\in {\cal E}\}$ as its neighbor set.
Each node is given a $m$-dimensional feature representation $\mathbf{x}_i$ and a $c$-dimensional one-hot class label $\mathbf{y}_i$. The feature inputs are then formed by
$\mathbf{X}=[\mathbf{x}_1, \cdots, \mathbf{x}_N]$, and the labels are $\mathbf{Y}=[\mathbf{y}_1, \cdots, \mathbf{y}_N]$.
Given the labels ${\bf Y}_{\cal L}$ of a subset of nodes ${\cal L} \subset {\cal V}$, the task of semi-supervised node classification is to predict the labels ${\bf Y}_{\cal U}$ of the unlabeled nodes ${\cal U} = {\cal V} \setminus {\cal L}$ by exploiting the graph structure ${\cal E}$ and the features of all the nodes $\mathbf{X}$. 

\subsection{Graph Neural Networks}
From a probabilistic view, most GNNs assume the local Markov property on node features, i.e., for each node $v_i$, 
the label $\mathbf{y}_i$ only depends on the node self-feature $x_i$ and its neighbor-features {$\mathbf{x}_j:j\in \mathcal{N}(v_i)$}.
We use superscript $l$ to indicate the layer index.
For the $l$-th layer of a GNN, we use $\mathbf{h}^l_i$ to represent the embedding of node $v_i$ and $\mathbf{h}^0_i$ to represent $\mathbf{x}_i$ or a projection of $\mathbf{x}_i$ for dimension reduction. 
Then, the general $l$-th layer $\operatorname{GraphConvolution}(\mathbf{A}, \mathbf{H}^{l-1})$ for node $i$ can be formulated as
\begin{align}
\mathbf{h}_{i}^{l}&=f\left(\mathbf{h}_{i}^{l-1},\left\{\mathbf{h}_{j}^{l-1}: j \in \mathcal{N}(v_i)\right\}\right),
\end{align}
where the graph convolution operator $f$ can be implemented by a weighted sum of each node based on the adjacent matrix ${\bf A}$ as in GCN~\cite{GCN} and GIN~\cite{xu2018powerful} or the attention mechanism in GAT~\cite{GAT}. The formulations of three well-known graph convolution layers are summarized in Table~\ref{tab:gnndetail}.

\begin{table}[t]
    \centering
\adjustbox{max width=8.6cm}{
\begin{tabular}{c|l}
\toprule
Models & \multicolumn{1}{c}{Aggregation for each layer $l(1 \leq l \leq L)$} \\ \midrule
$\operatorname{GCN}$ & $h_{v}^{(l)}=\sigma\left(\sum_{v^{\prime} \in \mathcal{N}_{v} \cup\{v\}} \frac{1}{\sqrt{\left(\left|\mathcal{N}_{v}\right|+1\right) \cdot\left(\left|\mathcal{N}_{v^{\prime}}\right|+1\right)}} \cdot W^{(l-1)} \cdot h_{v^{\prime}}^{(l-1)}\right)$ \\ \midrule
$\operatorname{GIN}$ & $ h_{v}^{(l)}=\mathrm{MLP}^{(l)}\left(\left(1+\epsilon^{(l)}\right) \cdot h_{v}^{(l-1)}+\sum_{{v^{\prime}} \in \mathcal{N}(v)} h_{v^{\prime}}^{(l-1)}\right)$ \\ \midrule
$\operatorname{GAT}$ & $ h_{v_{i}}^{(l)}=\sigma\left(\sum_{v_{j} \in \mathcal{N}_{v_{i}} \cup\left\{v_{i}\right\}} a_{i, j}^{(l-1)} \cdot W^{(l-1)} \cdot h_{v_{j}}^{(l-1)}\right)$ \\ \bottomrule

\end{tabular}}
    \caption{The different neighborhood aggregation schemes. Here $\sigma$ denotes the ReLu activation, $a$ denotes attention weights, $W$ denotes the weight matrix, and MLP denotes multiple layer perceptron.}   
    \label{tab:gnndetail}
\end{table}

The final output $\mathbf{Z} \in \mathbb{R}^{N \times c}$ of the label prediction is evaluated using a \textit{softmax} function to embed the last layer $\mathbf{H}^L$. The objective function is the cross-entropy of the ground truth labels $\mathbf{Y}$ and the output of the network $\mathbf{Z}$:
\begin{equation}
    \mathcal{O} = -\sum_{i \in \cal L} \sum_{j=1}^{c} \mathbf{Y}_{i j} \ln \mathbf{Z}_{i j}.
\end{equation}

\subsection{Homophily/Heterophily Metrics on Graphs}\label{sec:metric}
The homophily ratio $h$ aims to measure the overall homophily level in a graph. The commonly used node-level~\cite{pei2019geom} and edge-level~\cite{zhu2020beyond} homophily metrics are usually defined by
\begin{align}
\mathcal{H}_{\text {node }}(\mathcal{G})&=\frac{1}{|\mathcal{V}|} \sum_{v \in \mathcal{V}} \frac{\left|\left\{u \mid u \in \mathcal{N}_{v}, Y_u=Y_v\right\}\right|}{d_{v}}, \\
\mathcal{H}_{\text {edge }}(\mathcal{G})&=\frac{\left|\left\{e_{u v} \mid e_{u v} \in \mathcal{E}, Y_u=Y_v\right\}\right|}{|\mathcal{E}|}.
\end{align}
Such metrics measure the proportion of inter-class edges in a graph based on label consistency. {There is another metric $\mathcal{H}_{\text {agg }}(\mathcal{G})$ to consider the node features similarity from post-aggregation perspective~\cite{luan2021heterophily}. However, it is limited to the scenario where nodes have features.} By definition, the ratio $h \in [0, 1]$, graphs with $h$ closer to 1 tend to have more intra-class edges indicating stronger homophily; on the other hand, graphs with $h$ closer to 0 have more edges connecting different classes, which indicates stronger heterophily.

However, as reported in previous literature~\cite{zhu2020beyond,pei2019geom,luan2021heterophily}, these metrics are not significantly relevant to the prediction performance of GCNs. 
For example, for the well-known heterophily datasets Chameleon and Actor, their homo-ratios metrics $\mathcal{H}_\text{edge}$ are all 0.22. However, on the one hand, the reported accuracy of node classification for GCN varies, i.e., 60\% and 30\%, respectively~\cite{zhu2020beyond}. On the other hand, the accuracy of corresponding Multi-Layer Perceptrons(MLPs) is 46\% and 35\%. The opposite performance gap between GCN and MLP in the low homophilic datasets shows that inter-class edges can be either beneficial or harmful to classification, which motivates us to consider a new metric beyond the edge perspective to measure heterophily.

\section{{Review Heterophily from an Entire Neighbor Perspective}}\label{section:Metric}
\subsection{Measure the Graph-level Neighbor Effect for Heterophily}
In this section, we present a new metric to better understand the heterophily problem and answer whether all inter-class edges are harmful. As noted in Fig.~\ref{fig:intro}, 
when the label distribution of neighbors is random, i.e., every node is connected to other neighbors with random labels, there is no helpful information we can learn from the aggregation step. However, when the neighbor distribution of each class's nodes forms a certain identifiable distribution, regardless of whether the connected edges are intra- or inter-class, graph convolution can extract useful information from this non-random neighbor distribution for the downstream tasks. For instance, graph convolution can still achieve perfect 
performance on a bipartite graph~\cite{chien2020adaptive,ma2022is}.
Therefore, instead of simply calculating the proportion of inter-class edges as a measurement for graphs with heterophily, we need to measure the randomness/identifiability of the entire neighbor distribution.
\begin{figure}[t]
\begin{minipage}{1.0\linewidth}
  \centering
  \centerline{\includegraphics[width=8cm]{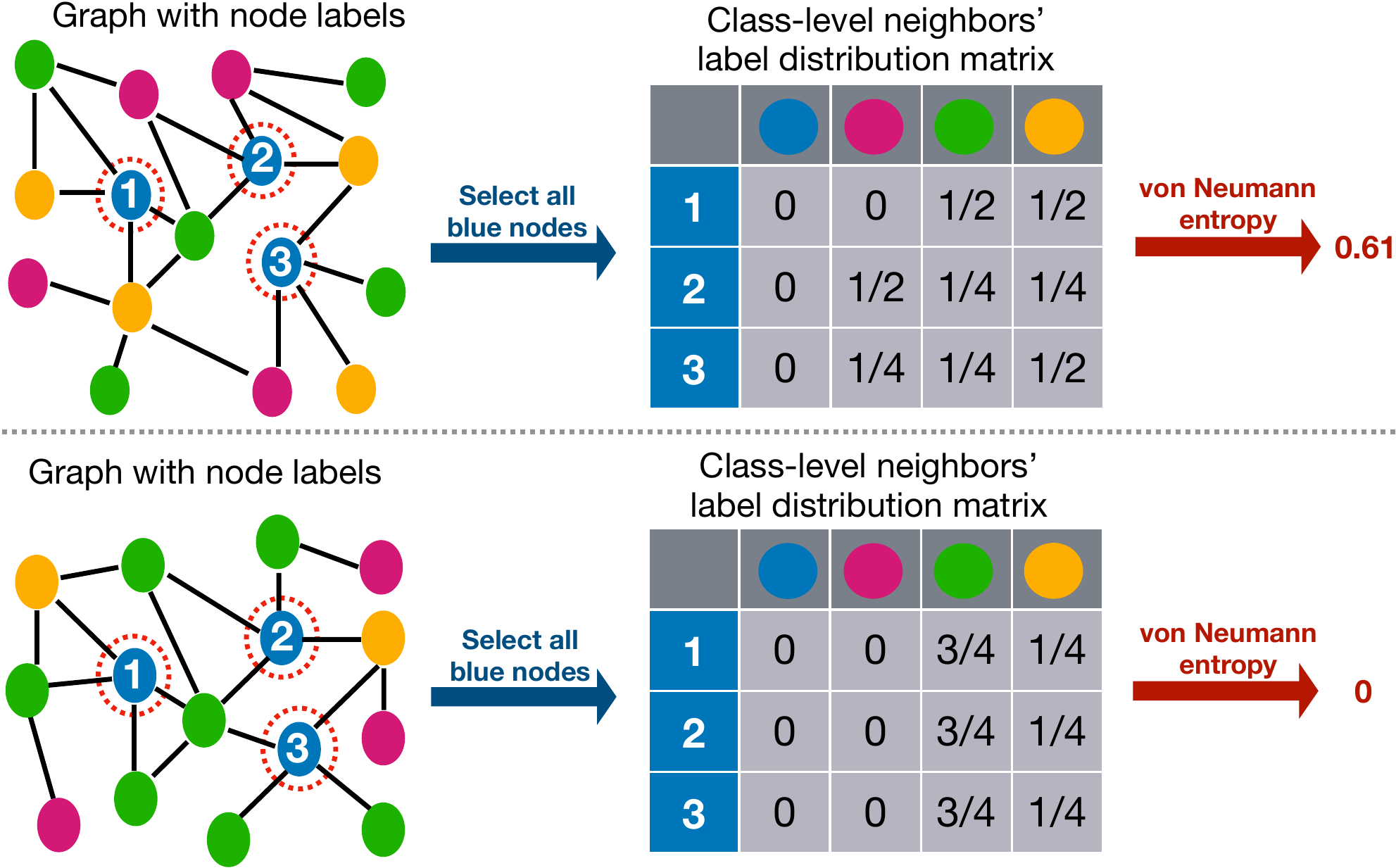}}
\end{minipage}
\caption{Class-level von Neumann entropy, which measures the information of neighbors' label distribution matrix. This metric ranges from $[0,1]$ and can quantify the \Q{identifiability} of neighbors for a specific class (a lower number indicates higher identifiability of neighbors).}
\label{fig:class-newMetric}
\end{figure}

We define the identifiability of neighbors as the information of the nodes' neighbor distribution, {which can be seen as the graph-level neighbor effect for heterophilic datasets}.
As Fig.~\ref{fig:class-newMetric} shows, to measure the identifiability of neighbors of each class, we group the nodes by class and form a class-level neighbor's label distribution matrix $A_{\mathcal N}^{k} \in \mathbb{R}^{n_{k} \times C}$ for each class $k$, where $k=1,...,C$ for different classes and $n_{k}$ indicates the number of nodes with label $k$.
Then, our task is to evaluate the information of the neighbors' label distribution matrix to quantify the identifiability of neighbors. 
Inspired by the von Neumann entropy in quantum statistical mechanics~\cite{bengtsson2017geometry}, 
{which measures the pureness/information of a quantum-mechanical system by calculating the entropy of the eigenvalue distribution of a positive definite symmetric density matrix}, 
we generalize this idea to our task of evaluating neighbors' identifiability.
{Specifically, since the neighbor distribution matrix $A_{\mathcal N}^{k}$ is not symmetric, we consider the entropy of the singular values distribution rather than the eigenvalues as an indicator of identifiability. This can be understood as a measurement of how many vectors (patterns of neighbors) are needed for an adequate explanation of the neighbors' label distribution matrix, indicating the richness/randomness of the neighbor distribution.}

%Specifically, 
Suppose $\sigma_1^k,\sigma_2^k, ...,\sigma_C^k $ denote singular values of $A_{\mathcal N}^{k}$, 
we then normalize them so that $\sum_{i=1}^{C}\sigma_i^k=1$, where $i=1,...,C$ for index of singular values.
Then we calculate the entropy of class K by
\begin{align}
\mathcal{H}_\text{neighbor}^{k}=\frac{-\sum_{i=1}^{C}\sigma_i^k \log(\sigma_i^k) }{\log(C)}.
\end{align}
The above metric ranges from $[0,1]$ and can be used to quantify the identifiability of neighbors for a specific class (the lower, the more identifiable). Considering the problem of class imbalance, we compute the weighted sum of class-level entropy to evaluate the %entropy of a graph
neighbors' identifiability of a graph.
\begin{align}
\mathcal{H}_\text{neighbor}(\mathcal{G}) = \sum_{k=1}^{C} \frac{n_{k}}{N}\mathcal{H}_\text{neighbor}^{k}.
\end{align}
\begin{figure}[t]
\begin{minipage}{1.0\linewidth}
  \centering
  \centerline{\includegraphics[width=8.8cm]
  {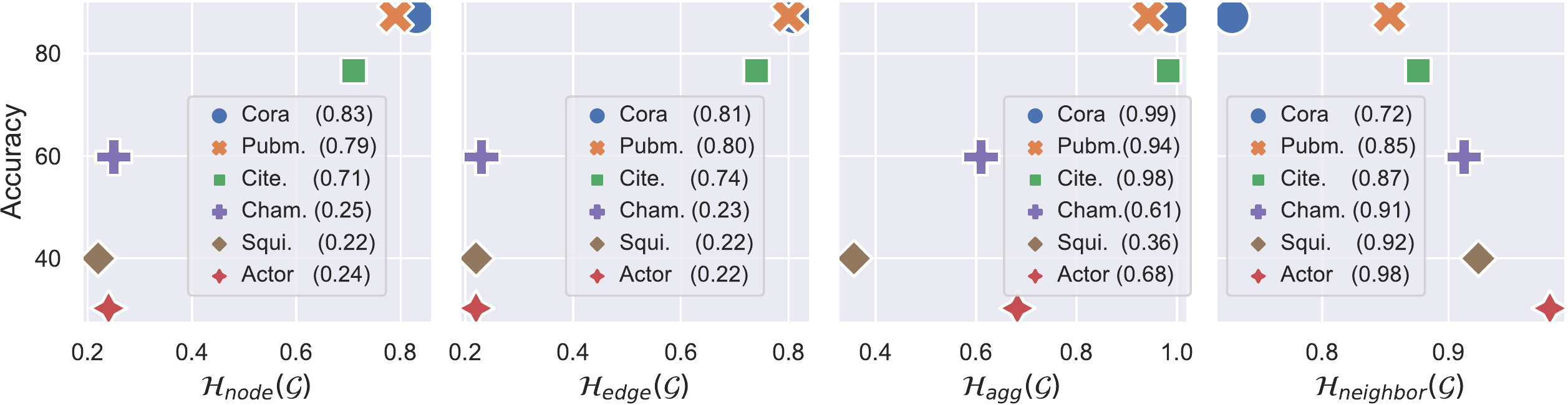}
  }
\end{minipage}
\caption{The relation between hom-metric and GCN's performance from \protect\cite{zhu2020beyond} on different graph node classification datasets\protect\footnotemark. %Our metric are more monotonous with the models' performance.
Our metric $\mathcal{H}_{\text{neighbor}}(\mathcal{G})$ is more monotonous with the models' performance. Moreover, compared with other metrics, it can distinguish the Actor dataset, which has a random neighbor distribution and the lowest accuracy. More details can be found in Section~\protect\ref{metrics_vs_performance}.}
\label{fig:dataset-newMetric}
\end{figure}
\footnotetext{We ignore the datasets that have fewer than 500 nodes. Their performances are highly sensitive to the data splits.}

Our metric sheds new light on understanding the heterophily problem from an identifiable neighbor perspective. As shown in Fig.~\ref{fig:dataset-newMetric}, compared with the node/edge-level homophily metrics, our measurement and the GCNs' performance for different datasets are more monotonous. 
Specifically, we can observe that the Actor, Chameleon, and Squirrel datasets have similar node/edge-level homophily metrics (proportion of inter-class edges) in the {first} and {second} parts of Fig.~\ref{fig:dataset-newMetric}. However, the performance of these datasets is inconsistent, especially for the Actor dataset, which has the poorest performance (accuracy$\approx$30\%) and is totally different from the others. {As we can see in the third part of Fig.~\ref{fig:dataset-newMetric}, the $\mathcal{H}_{\text{agg}}(\mathcal{G})$ still fail to distinguish the Actor from others.} In contrast, as shown in the right of Fig.~\ref{fig:dataset-newMetric}, our metric can distinguish the Actor dataset from others since it has a nearly random neighbor distribution ($\mathcal{H}_\text{neighbor}$=0.98).
This indicates that the proposed metric can better explain the difference in model performance from a neighbor perspective. 
Furthermore, our metric reveals that the inter-class edges are not always harmful for the node classification during aggregation, and the entire local neighbor perspective can provide more information. 

\subsection{Improve GNNs via Learning Node-level Neighbor Effect}
Unlike other specific heterophilic-oriented GNNs that need to modify graph convolution kernels, our neighbor perspective motivates us to learn the effectiveness of each node's local neighbors during aggregation to help traditional GNNs deal with heterophily in a general way. 
Most GNNs straightforwardly feed the current aggregation features to the following graph convolution layer and adopt the last layer representation for the downstream node classification task~\cite{GCN,GAT}. 
However, the entanglement of aggregation and classification may lead to over-smoothing of node representations due to inter-class neighbors, resulting in a loss of discrimination in heterophilic graphs, despite the fact that neighbor information may be useful for downstream node classification.
To adaptively combine neighbor information and enhance traditional GNNs, we propose decoupling discriminant representations of nodes from the aggregation. Then, it allows us to guide the aggregation and generate suitable representations of nodes for classification by learning the node-level neighbor effect.

However, to guide the aggregation during the training process, we cannot directly utilize the entropy measurement and need to evaluate the node-level neighbor effect in another way. The reasons are twofold: (1) Similar to $\mathcal{H}_\text{edge}$ and $\mathcal{H}_\text{node}$, the computation of the entropy $\mathcal{H}_\text{neighbor}$ also requires the labels of all the nodes, which are unavailable in the training process. (2) The class-level neighbor distribution identifiability $\mathcal{H}_\text{neighbor}^c$ is not consistent with the entropy of the node-level label distribution. Namely, the entropy of a node does not represent the identifiability of the neighbor distribution of this class. 
In the next section, we will elaborate on how to adaptively learn the node-level neighbor effect from the downstream supervision signal and the features of a node with its neighbor.

\section{Proposed Method}\label{Sec:5}
In this section, we propose the Conv-Agnostic GNN framework ({\ourModelName}) to improve traditional GNNs performance by adaptively learning the node-level neighbor effect. Then, we provide a spectral analysis to show {the expressive power of the proposed framework on the node classification task}. Finally, we describe the difference between our framework and the GNNs with decoupling design and skip connection.

\subsection{{Conv-Agnostic GNN Framework}}
\begin{figure}[t]
%\begin{minipage}[b]{1.0\linewidth}
  \centering
  \centerline{\includegraphics[width=\linewidth]{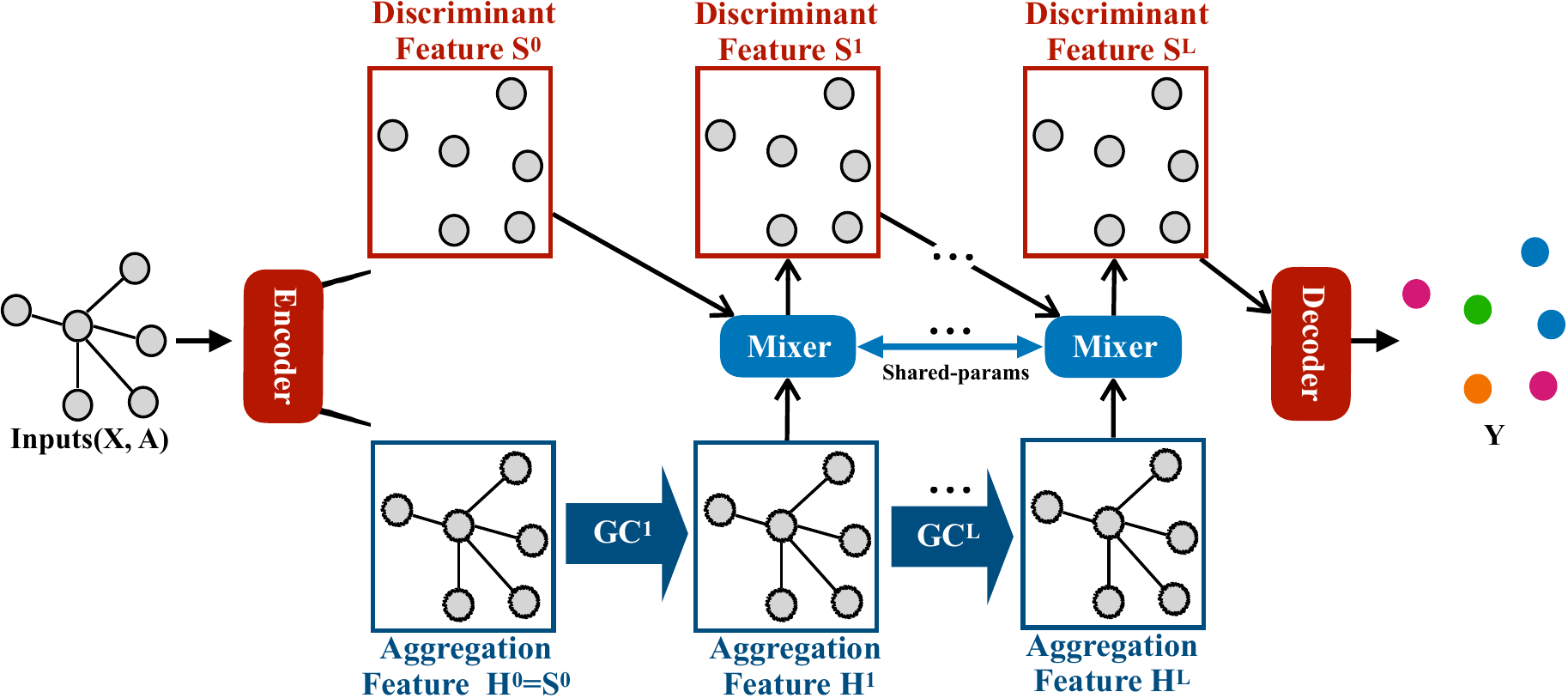}}
%\end{minipage}
\caption{The GC operator indicates any Graph Convolutions. The shared-parameter Mixer function can help each node determine whether to absorb neighbors' information by considering the \Q{neighbor effect} based on the node's feature.}
\label{fig:framework}
\end{figure}

The proposed \ourModelName aims to empower traditional GNNs to generate suitable representations for each node for both homophilic and heterophilic graphs. 
{The core idea is to treat the node-local neighbor as an entirety and determine its efficacy during aggregation. To this end, we first decouple the representation of nodes into discrimination and aggregation.} Then, we learn a mixer module that can adaptively evaluate each node's neighbor effect based on these two representations and determine whether to incorporate the information from neighbors.
As shown in Fig.~\ref{fig:framework}, our framework is composed of four major components: {\it Encoder}, {\it Graph Convolution (GC)}, {\it Mixer}, and {\it Decoder}. 
Below, we elaborate on each component in order. Moreover, we append the normalization operation after the {\it Encoder}, {\it GC}, and {\it Mixer} to maintain the numerical stability, which will be discussed later.

\vspace{1pt}
\noindent
\textbf{Encoder}: We use a linear layer as the encoder to transform the node features $\mathbf{X}$. Then, we feed it into two streams. One is the node's own feature {${\bf S}^0 \in \mathbb{R}^{N \times d}$} for downstream-task discrimination, and the other is the aggregation feature {${\bf H}^0 \in \mathbb{R}^{N \times d}$} for the graph convolution. 
These two streams decouple the discriminant feature and the information from neighbors, 
which may prevent representations of nodes for classification from being over-smooth by their neighbors during aggregation.
Moreover, this decoupling operation allows each node to evaluate the neighbor effect for downstream discrimination tasks via the mixer module in the following aggregation step.
\begin{align}
\mathbf{H}^{0} = \mathbf{S}^{0} =  \operatorname{Norm}(f_{\operatorname{{encoder}}}(\mathbf{X})).
\end{align}

\vspace{1pt}
\noindent
\textbf{Graph Convolution (GC)}: Since our framework is Conv-Agnostic, in this part, any standard graph convolution layers (e.g., GCN, GAT, and GIN) can be applied to aggregate each node's neighborhood information to update the aggregation representation $\mathbf{H}$. 
\begin{align}
\mathbf{H}^{l+1} = \operatorname{Norm}(\operatorname{GraphConvolution}(\mathbf{A}, \mathbf{H}^l)).
\end{align}
Moreover, this graph convolution layer can be stacked multiple times to enhance the receptive field of each node by considering the information of more neighbors. The embedding $\mathbf{H}^l$ can also be regarded as the $l$-hop neighbors' information. 
However, most GNNs ignore the fact that the information from neighbors is not always beneficial for the classification task of each node. Hence, we propose the following mixer operator to determine the neighbor effect of each hop.

\vspace{1pt}
\noindent
\textbf{Mixer}: From the view of node $v_i$ at layer $l$, it needs to combine the discrimination feature $\mathbf{S}_i^l$ and $l$-hop neighbors feature $\mathbf{H}_i^l$ according to the \Q{neighbor effect} to update the representative embedding for the downstream task. Therefore, the goal of the mixer function is to evaluate the \Q{neighbor effect} of each node and then to selectively incorporate the neighbors' information. 
As discussed in Section~\ref{section:Metric}, we do not need to inject the entropy function to evaluate the node-level neighbor effect. We have already encoded the neighbor information of each node into $\mathbf{H}^l$ by the graph convolution. According to the universal approximation theorem~\cite{hornik1989multilayer,hornik1991approximation},  we hypothesize that the MLP can adaptively learn the node-level neighbor effect based on {$\mathbf{S}^{l-1} \in \mathbb{R}^{N \times d}$} and {$\mathbf{H}^l \in \mathbb{R}^{N \times d}$} and the downstream objective $\mathcal{O}$. For simplicity, we implement the mixer by
\begin{align}\label{eqn::mixer}
{\bm{\alpha}}^l &= \sigma (f_{\operatorname{{mixer}}}(\mathbf{S}^{l-1} \parallel \mathbf{H}^l)),\\
\mathbf{S}^{l}&=\operatorname{Norm}(({\bf 1}-{\bm{\alpha}}^l) * \mathbf{S}^{l-1} + {\bm{\alpha}}^l * \mathbf{H}^{l}),
\end{align}
{where $\parallel$ is the concat operator and the function $f_{\operatorname{{mixer}}}$ is a linear layer $ \mathbb{R}^{N \times 2d}\rightarrow \mathbb{R}^{N \times 1}$}. It maps the discrimination feature $\mathbf{S}^{l-1}$ and neighbor feature $\mathbf{H}^{l-1}$ at layer $l$ to a vector ${\bm{\alpha}}^l \in \mathbb{R}^{N \times 1}$. With the sigmoid function $\sigma$, each element of $\bm{\alpha}^l$ is normalized to an importance score ranging from 0 to 1 and can be regarded as the neighbor effect of the node.
Based on the \Q{importance score} $\bm{\alpha}^l$, we use the convex combination of discriminant feature $\mathbf{S}^{l-1}$ and neighbor information $\mathbf{H}^{l-1}$ at each layer to adaptively update the discriminant feature $\mathbf{S}^{l}$. 
This convex combination strategy has the advantages of numerical stability and interpretability and has been widely used in modern deep learning, such as the Highway network~\cite{srivastava2015highway} and Attention mechanism~\cite{vaswani2017attention}.

Note that the parameters of all the $f_{\operatorname{{mixer}}}$ functions are shared across layers to learn the \Q{neighbor effect} of each node. {Such a parameter-efficient sharing mechanism can help reduce overfitting and improve generalization, similar to the gating function in Long Short-Term Memory~\cite{hochreiter1997long}.} Moreover, the mixer function maintains the expressive power of the chosen graph convolution, since it can easily degenerate to the normal graph convolution when $\bm{\alpha}^l={\bf 1}$. \Q{As shown in the ablation study, our implementation of this mixer function is very effective with minimal parameter cost.}

\vspace{4pt}
\noindent
\textbf{Decoder}: 
With the last layer discriminant $\mathbf{S}^L$ at hand, the task of the decoder is to produce the final prediction $\mathbf{Z}$ for classification. 
For simplicity, we use the linear layer with {\it softmax} operator as our decoder $f_{\operatorname{{decoder}}}$. 
\begin{align}
\mathbf{Z} = \operatorname{softmax}(f_{\operatorname{{decoder}}}(\mathbf{S}^L)).
\end{align}

Note that we also apply the Norm layer to maintain numerical stability at the end of the {\it Encoder}, {\it GC} and {\it Mixer} modules. Specifically, we use L2 normalization for each node. Compared with the widely used BatchNorm and LayerNorm, we experimentally show that the parameter-free L2 normalization can achieve better performance for our task.

{In summary, unlike other GNNs that explicitly modify graph filters or graph convolution kernels for heterophily~\cite{bo2021beyond,chien2020adaptive,luan2021heterophily}, our node-level mixer is a convolution-agnostic technique. We divide the representation of the node into two parts: node-self for discrimination and neighbor information from arbitrary graph convolution. Hence the original graph convolution operator may not directly interact with the node's discrimination feature and over-smooth its representation. Moreover, we use only one shared-parameterized layer to adaptively learn node-level neighbor effect and mix these two representations, which can be seen as adding a plug-in side layer to standard GNNs for generating more discriminative node representations. Therefore, our method is compatible with most graph convolution layers (GCN, GAT, GIN, etc.) and enhances their performance in a more general and parameter-efficient way. 
}

\subsection{{Spectral Analysis of CAGNN}}
We then study the expressive power of \ourModelName 
from a spectral perspective. 
Recall the $K$-order polynomial graph filters with graph signal $\mathbf{X}$ and propagation matrix $\mathbf{P}$ as $\textstyle{\left(\sum_{l=0}^{K}{\theta}^{l} {\mathbf{P}}^{l}\right) \mathbf{X}}$, where $\textstyle{\mathbf{P}=\mathbf{D}^{-1/2}\mathbf{A}\mathbf{D}^{-1/2}}$ is the normalized adjacency matrix and the scalar $\theta$s are the polynomial coefficients.
Note that using such a polynomial graph filter can derive either high- or low-pass filters~\cite{shuman2013emerging,he2021bernnet},
maintaining the model's ability to deal with various label-connected patterns, which is essential to prevent over-smoothing~\cite{chen2020simple} and to learn from heterophilic graphs~\cite{chien2020adaptive}.
However, most spectral GNNs learn the shared $K$-order polynomial scalar coefficients $\theta$'s for all nodes, which may limit the expressive power.
We demonstrate that \ourModelName corresponds to a polynomial graph filter with different coefficients for each node on the graph spectral domain. 
\begin{theorem}
\textit{
Considering the propagation matrix $\mathbf{P}$ used for the basic graph convolution layer and a graph signal $\mathbf{X}$, a $K$-layer \ourModelName has the ability to express a $K$-order polynomial filter $\left(\sum_{l=0}^{K} \bm{\theta}^{l} {\mathbf{P}}^{l}\right) \mathbf{X}$ with different coefficients ${\theta}$s for each node.}
\label{theorem:th.1}
\end{theorem}

The above theorem indicates the expressive power on node classification of our framework.
Intuitively, the importance score ${\bm{\alpha}}$s evaluated by the {\it Mixer} allows \ourModelName to simulate different coefficient ${{\theta}}$s of the polynomial graph filter for each node. Note that with a proper choice of ${\theta}$s, the discriminant feature $\mathbf{S}^K$ of each node can carry information from both the input feature and the high-order neighbor's information adaptively with the increment of the order $K$. Compared with other spectral GNNs that learn shared polynomial filters $\textstyle{\left(\sum_{l=0}^{K}{\theta}^{l} {\mathbf{P}}^{l}\right) \mathbf{X}}$ for all nodes~\cite{defferrard2016convolutional,chien2020adaptive,chen2020simple}, our models can empower each node with distinct polynomial coefficients by vectorized $\bm{\theta}^l$. 
This property can capture the nodes' distinct complex connected patterns, as in Section~\ref{classification-comparison}, our framework achieves higher performance than other spectral GNNs experimentally.
The detailed proof is presented below.

\textit{Proof. }For simplicity, we neglect the L2 normalization at each layer because the simplified version also produces comparable performance. Moreover, we assume the input feature $\mathbf{X}$ to be non-negative and remove the nonlinear ReLU activation in the graph convolution layer~\cite{chen2020simple,wu2019simplifying}.
Then, for the simplified graph convolution layer, we have
\begin{align}
\mathbf{H}^{l}&=\mathbf{P}\mathbf{H}^{l-1}\mathbf{W}^{l}\\
&=\mathbf{P}^{l}\mathbf{H}^0  {\textstyle \prod_{j=0}^{l}}\mathbf{W}^{j}\\ 
&=\mathbf{P}^{l}\mathbf{X} \mathbf{\tilde{W}}^{l}, \quad \text{where } \mathbf{\tilde{W}}^{l}={\textstyle \prod_{j=0}^{l}}\mathbf{W}^{j}.
\end{align}
Furthermore, we can express the K-layer representation as
%{\mathbf{S}}^{0} &= {\mathbf{H}}^{0} \\
\begin{align}
{\mathbf{S}}^{K} &= ({\bf 1}- {\bm{\alpha}}^{K})*{\mathbf{S}}^{K-1} +  {\bm{\alpha}}^{K} * {\mathbf{H}}^K \\
&=\sum_{l=0}^{K}  {\bm{\alpha}}^{l} {\textstyle \prod_{k=l+1}^{K}} (1- {\bm{\alpha}}^{k})*{\mathbf{H}}^{l} \\
 &=\sum_{l=0}^{K}  {\bm{\vartheta}}^{l} * \mathbf{P}^{l}\mathbf{X}\mathbf{\tilde{W} }^{l}, 
 %\text{where }  {\bm{\vartheta}}^{l}= {\bm{\alpha}}^{l} {\textstyle \prod_{k=l+1}^{K}} (1- {\bm{\alpha}}^{k}).
  \text{where }  {\bm{\vartheta}}^{l}= {\bm{\alpha}}^{l} {\textstyle \prod_{k=l+1}^{K}} (1- {\bm{\alpha}}^{k}).
\end{align}
However, ${\bm{\vartheta}}^{l} \in [0,1]^{N \times 1}$ which limits the expressive power of the polynomial filter. Thanks to the weight matrix $\mathbf{\tilde{W} }^{l}$, the coefficients of the polynomial can be extended to arbitrary values. Inspired by~\cite{chen2020simple}, we consider a weaker version of \ourModelName by fixing the weight matrix $\mathbf{\tilde{W} }^{l}$ to be $\bm{\gamma}^{l}$, where $\bm{\gamma}^{l}$ is a learnable parameter. We have
\begin{align}
{\mathbf{S}}^{K} &=\sum_{l=0}^{K}  {\bm{\vartheta}}^{l} * \mathbf{P}^{l}\mathbf{X}{\bm{\gamma}}^{l} \\
&= \sum_{l=0}^{K}  {\bm{\theta }}^{l} * \mathbf{P}^{l}\mathbf{X},\quad  \text{where }  {\bm{\theta}}^{l}= {\bm{\vartheta}}^{l}{\bm{\gamma} }^{l}.
\end{align}
The polynomial coefficient ${\bm{\theta}}^{l}$ for each layer $l$ can be set to desired values with the help of the scalable parameter $\bm{\gamma}^{l}$, which concludes the proof.

\subsection{Relation to other Decouple or Skip Connection GNNs}
In this section, we discuss the relation of our two techniques (decoupled operation and mixer module) in the framework with other models.

\vspace{4pt}
\noindent
\textbf{Relation to GNNs with Decouple Design}: We remark that the decoupling operation, which we study in this work, is a distinct concept from other decoupling GNNs. Well-known decoupled GNNs, such as APPNP~\cite{klicpera2018predict},
S$^2$GC~\cite{zhu2021simple},
and GPRGNN~\cite{chien2020adaptive}, aim to decouple the propagation and transformation of graph convolution. Without the loss of generalizability, they can be formulated as follows.
\begin{align}
 \mathbf{H}^K &= \underbrace{(\sum_{l=0}^K \theta^l\mathbf{P}^l)}_{propagate} \overbrace{\mathbf{H}^0\mathbf{W}}^{transform},  
\end{align}
where ${\theta^l}$ is a scalar and $\mathbf{P}$ is the propagation matrix. This operation makes the node have the ability to receive high-order neighbor information in one layer. In addition, the learnable parameter $\theta^l$ ensures that it can learn the arbitrary coefficient of polynomial graph filters beyond the low-pass filters and perform well in heterophily~\cite{chien2020adaptive}. However, it needs to decouple the weight matrix $\mathbf{W}$ to reformulate the graph convolution operator, which restricts the adaptation to other graph convolutions.

In contrast, our approach aims to decouple the discriminant feature and the aggregation feature, which can be easily compatible with most standard GNNs. Moreover, we theoretically prove that our framework can learn different arbitrary coefficients of K-order polynomial graph filters for each node and achieve higher performance experimentally.

\vspace{4pt}
\noindent
\textbf{Relation to GNNs with Skip Connection}:
To better understand the mixer module, we compare it with respect to the DeepGNNs with skip connections. Note that we aim at the heterophily problem, and our approach is different from the DeepGNNs designed to reduce the over-smoothing problem~\cite{li2018deeper}. 
To alleviate the over-smoothing problem when the model becomes deeper, most DeepGNNs equip the residual connection or initial connection to combine previous layers' features to prevent forgetting the original feature when models become deeper~\cite{chen2020simple,li2019deepgcns}. The standard DeepGNN with skip connection can be formulated as follows.
\begin{align}
\mathbf{H}^l=\sigma(\mathbf{P}\mathbf{H}^l\mathbf{W}^l+\left \{\mathbf{H}^{l-1}/ \mathbf{H}^0\right \})\text{(residual/init connection)}.
\end{align}
However, the DeepGNNs with skip connections are not designed for heterophilic graphs. Most skip connections combine previous features and cannot adaptively aggregate information from neighbors for each node.
For instance, the recent well-known SOTA of this type model is GCNII~\cite{chen2020simple}, where
\begin{align}
\mathbf{H}^{l}=\sigma\left(\left(\left(1-\alpha^{l}\right) {\mathbf{P}} \mathbf{H}^{l}+\alpha^{l} \mathbf{H}^{0}\right)\left(\left(1-\beta^{l}\right) \mathbf{I}+\beta^{l} \mathbf{W}^{l}\right)\right). 
\end{align}
The scalar $\alpha^l$ in GCNII is a manually chosen hyper-parameter that controls the strength of initial connection. However, since the $\alpha^l$ is shared for all nodes and not learnable, it is not the best choice to guide how to aggregate neighbors' information on heterophilic datasets. 

In contrast, as shown in Equations~\eqref{eqn::mixer}, we use the mixer function to learn the importance score $\bm{\alpha}$ to explicitly evaluate the node-level neighbor effect for feature fusion.
Moreover, compared with the skip connection, we separate the discriminant features $\mathbf{S}$ of nodes at each layer and do not feed them into the next layer graph convolution component.

\section{Experiments}\label{Sec:6}
In this section, we report and compare the results for node classification on both real-world heterophily and homophily datasets to investigate the effectiveness, robustness and interpretability of the proposed heterophily GNN framework \ourModelName. We also show the importance of the neighbor effect from the relationship between the model performance and metrics. 

\setlength{\tabcolsep}{3pt}
\begin{table*}[htbp]
   \scriptsize
    \centering  
    \vspace{-0.35cm}
    \caption{Performance comparison on various real-world heterophily and homophily datasets. Mean test accuracy and standard deviation are reported over 10 random data splits. The best performance is highlighted. ``*" denotes the results obtained from~\protect\cite{zhu2020beyond}.}
    \label{tab:ssnc-results}
    \begin{adjustbox}{width=\textwidth}
    \begin{tabular}{l|cccccc|ccc|c} %
    \toprule
       & \texttt{\bf Texas}& \texttt{\bf Wisconsin}& \texttt{\bf Actor} & \texttt{\bf Squirrel} & \texttt{\bf Chameleon} &
       \texttt{\bf Cornell}& \texttt{\bf Citeseer} &   \texttt{\bf Pubmed} &   \texttt{\bf Cora} & {\textbf{Average}}\\
\midrule
\textbf{Nodes }  & {183} & {251} & {7,600} & {5,201} & {2,277} & {183} & {3,327} & {19,717} & {2,708} & - \\
\textbf{Edges }  & {295} & {466} & {26,752} & {198,493} & {31,421} & {280} & {4,676} & {44,327} & {5,278} & - \\
\textbf{Features }  & {1,703} & {1,703} & {931} & {2,089} & {2,325} & {1,703} & {3,703} & {500} & {1,433} & - \\
\textbf{Classes }  & {5} & {5} & {5} & {5} & {5} & {5} & {7} & {3} & {6} & -\\ \midrule
${\mathcal{H}_\text{node}}$ & {0.06} & {0.16} & {0.24} & {0.22} & {0.25} & {0.11} & {0.71} & {0.79} & {0.83} &  -\\
${\mathcal{H}_\text{edge}}$ & {0.11} & {0.21} & {0.22} & {0.22} & {0.23} & {0.30} & {0.74} & {0.80} & {0.81} &  -\\
${\mathcal{H}_\text{agg}}$ & {0.69} & {0.78} & {0.68} & {0.36} & {0.61} & {0.80} & {0.98} & {0.94} & {0.99} &  -\\
${\mathcal{H}_\text{neighbor}}$ & {0.45} & {0.72} & \textbf{0.98} & {0.92} & {0.91} & {0.55} & {0.87} & {0.85} & {0.72} & -\\
\midrule
    {GEOM-GCN\cite{pei2019geom}*} & $67.57$ & $64.12$ & $31.63$ & $38.14$ & $60.90$ & $60.81$ 
	   & \cellcolor{blue!15}${77.99}$ &  \cellcolor{blue!15}${90.05}$ & $85.27$ & 64.05 \\
	   {MixHop\cite{abu2019mixhop}*} & $77.84{\scriptstyle\pm7.73}$ & $75.88{\scriptstyle\pm4.90}$ & $32.22{\scriptstyle\pm2.34}$ & $43.80{\scriptstyle\pm1.48}$ & $60.50{\scriptstyle\pm2.53}$ & $73.51{\scriptstyle\pm6.34}$ & 
	   $76.26{\scriptstyle\pm1.33}$ & $85.31{\scriptstyle\pm0.61}$ & ${87.61}{\scriptstyle\pm0.85}$ & 68.21\\
       {H2GCN-1\cite{zhu2020beyond}*} & 
       $84.86{\scriptstyle\pm6.77}$ & 
       ${86.67}{\scriptstyle\pm4.69}$ & \cellcolor{blue!15}$35.86{\scriptstyle\pm1.03}$ & $36.42{\scriptstyle\pm1.89}$ & $57.11{\scriptstyle\pm1.58}$ & $82.16{\scriptstyle\pm4.80}$ & 
       $77.07{\scriptstyle\pm1.64}$ & $89.40{\scriptstyle\pm0.34}$ & $86.92{\scriptstyle\pm1.37}$ & 70.72\\
      {H2GCN-2\cite{zhu2020beyond}*} & 
      $82.16{\scriptstyle\pm5.28}$ & $85.88{\scriptstyle\pm4.22}$ & $35.62{\scriptstyle\pm1.30}$ & $37.90{\scriptstyle\pm2.02}$ & $59.39{\scriptstyle\pm1.98}$ & $82.16{\scriptstyle\pm6.00}$ & 
      $76.88{\scriptstyle\pm1.77}$ & $89.59{\scriptstyle\pm0.33}$ & $87.81{\scriptstyle\pm1.35}$ & 70.87\\
	   {GPRGNN\cite{chien2020adaptive}} & $82.12{\scriptstyle\pm7.72}$ & $81.16{\scriptstyle\pm3.17}$ & 
	   $33.29{\scriptstyle\pm1.39}$ & 
	   $43.29{\scriptstyle\pm1.66}$ & 
	   $61.82{\scriptstyle\pm2.39}$ &  
	   $81.08{\scriptstyle\pm6.59}$ &  
	   $75.56{\scriptstyle\pm1.62}$ & 
	   $86.85{\scriptstyle\pm0.46}$ & 
	   $86.98{\scriptstyle\pm1.33}$ & 70.15
	   \\ 
	   {JK-Net\cite{xu2018representation}*} & $66.49{\scriptstyle\pm6.64}$ & $74.31{\scriptstyle\pm6.43}$ & $34.18{\scriptstyle\pm0.85}$ & $40.45{\scriptstyle\pm1.61}$ & $63.42{\scriptstyle\pm2.00}$ & $64.59{\scriptstyle\pm8.68}$ & 
	   $74.51{\scriptstyle\pm1.75}$ & $88.41{\scriptstyle\pm0.45}$ & $86.79{\scriptstyle\pm0.92}$ & 65.79\\
	   {GCN-Cheby\cite{defferrard2016convolutional}*} & $77.30{\scriptstyle\pm4.07}$ & $79.41{\scriptstyle\pm4.46}$ & $34.11{\scriptstyle\pm1.09}$ & $43.86{\scriptstyle\pm1.64}$ & $60.50{\scriptstyle\pm2.53}$ & $73.51{\scriptstyle\pm6.34}$ & 
	   $76.26{\scriptstyle\pm1.33}$ & $85.31{\scriptstyle\pm0.61}$ & $87.61{\scriptstyle\pm0.85}$ & 68.65\\
	   {FAGCN\cite{bo2021beyond}} & $78.11 {\scriptstyle\pm 5.01}$ & $81.56{\scriptstyle\pm4.64}$ & $35.41{\scriptstyle\pm1.18}$ & $42.43{\scriptstyle\pm2.11}$ & $56.31{\scriptstyle\pm3.22}$ & $76.12{\scriptstyle\pm7.65}$ & 
	   $74.86{\scriptstyle\pm2.42}$ & 
	   $85.74{\scriptstyle\pm0.36}$ & $83.21{\scriptstyle\pm 2.04}$ & 68.18 
    \\
	   {GCNII\cite{chen2020simple}} & 69.72 ${\scriptstyle\pm8.90}$ & $75.29{\scriptstyle\pm4.64}$ & $35.58{\scriptstyle\pm1.25}$ & $47.21{\scriptstyle\pm1.73}$ & $60.79{\scriptstyle\pm2.35}$ & $79.19{\scriptstyle\pm6.12}$ & $76.82{\scriptstyle\pm1.67}$ & $89.26{\scriptstyle\pm0.48}$ & \cellcolor{blue!15}$87.89{\scriptstyle\pm1.88}$ & 69.07 \\
	   {MLP*} & $81.89{\scriptstyle\pm4.78}$ & $85.29{\scriptstyle\pm3.61}$ & $35.76{\scriptstyle\pm0.98}$ & $29.68{\scriptstyle\pm1.81}$ & $46.36{\scriptstyle\pm2.52}$ & $81.08{\scriptstyle\pm6.37}$ & 
	   $72.41{\scriptstyle\pm2.18}$ & $86.65{\scriptstyle\pm0.35}$ & $74.75{\scriptstyle\pm2.22}$ &  65.99\\
       \midrule
	   {GIN\cite{xu2018powerful}}& $71.89{\scriptstyle\pm6.64}$ & $77.84{\scriptstyle\pm4.53}$ &
	   $32.15{\scriptstyle\pm1.56}$ & 
	   $35.78{\scriptstyle\pm1.34}$ & 
	   $56.18{\scriptstyle\pm1.94}$ & 
	   $75.94{\scriptstyle\pm7.47}$ & 
	   $75.68{\scriptstyle\pm1.89}$ & 
	   $88.64{\scriptstyle\pm0.54}$ & 
	   $86.49{\scriptstyle\pm1.91}$ & 66.73\\
	   \textbf{$\text{CAGNN}_{GIN}$} & $82.51{\scriptstyle\pm4.49}$ &
	   $84.41{\scriptstyle\pm2.98}$ & 
	   $35.09{\scriptstyle\pm1.07}$ & 
	   $54.07{\scriptstyle\pm1.39}$ & 
	   $67.32{\scriptstyle\pm1.95}$ & 
	   \cellcolor{blue!15}$82.97{\scriptstyle\pm5.56}$ & 
	   $76.64{\scriptstyle\pm1.27}$ & 
	   $89.46{\scriptstyle\pm0.48}$ &
	   $87.06{\scriptstyle\pm1.02}$ & 73.28\\
	   {$Gain$}& $\uparrow 14.77\%$ & $\uparrow 8.44\%$ & $\uparrow 9.14\%$ & $\uparrow 51.12\%$ & $\uparrow 19.83\%$ & $\uparrow 9.26\%$ 
	   & $\uparrow 1.21\%$ & $\uparrow 0.93\%$ & $\uparrow 0.66\%$ & $\uparrow 9.81\%$\\
       \midrule
	   {GAT\cite{GAT}*} & $58.38{\scriptstyle\pm4.45}$ & $55.29{\scriptstyle\pm8.71}$ & $26.28{\scriptstyle\pm1.73}$ & $30.62{\scriptstyle\pm2.11}$ & $54.69{\scriptstyle\pm1.95}$ & $58.92{\scriptstyle\pm3.32}$ & 
	   $75.46{\scriptstyle\pm1.72}$ & $84.68{\scriptstyle\pm0.44}$ & $82.68{\scriptstyle\pm1.80}$ & 58.56 \\
	   \textbf{$\text{CAGNN}_{GAT}$} &
	   $83.52{\scriptstyle\pm6.17}$ & 
	   \cellcolor{blue!15}$86.68{\scriptstyle\pm3.74}$ &
	   $34.95{\scriptstyle\pm1.36}$ & 
	   $55.52{\scriptstyle\pm1.45}$ & 
	   $68.50{\scriptstyle\pm1.32}$ & 
	   $81.35{\scriptstyle\pm5.28}$ & 
	   $75.51{\scriptstyle\pm1.54}$ & 
	   $89.51{\scriptstyle\pm0.54}$ & 
	   $87.48{\scriptstyle\pm1.10}$ & 73.67 \\
	   
	  {$Gain$}  & $\uparrow 43.06\%$ & $\uparrow 56.77\%$ & $\uparrow 32.99\%$ & $\uparrow 81.32\%$ & $\uparrow 25.27\%$ & $\uparrow 38.07\%$ 
	   & $\uparrow 0.07\%$ & $\uparrow 5.70\%$ & $\uparrow 5.81\%$ & $\uparrow 25.81\%$\\
       \midrule
	   {GCN\cite{GCN}*} & $59.46{\scriptstyle\pm5.25}$ & $59.80{\scriptstyle\pm6.99}$ & $30.26{\scriptstyle\pm0.79}$ & $36.89{\scriptstyle\pm1.34}$ & $59.82{\scriptstyle\pm2.58}$ & $57.03{\scriptstyle\pm4.67}$ &  
	   $76.68{\scriptstyle\pm1.64}$ & $87.38{\scriptstyle\pm0.66}$ & $87.28{\scriptstyle\pm1.26}$ &  61.62\\
	   \textbf{$\text{CAGNN}_{GCN}$} & 
	   \cellcolor{blue!15} $85.13{\scriptstyle\pm 5.73}$ & $82.55{\scriptstyle\pm 4.17}$ & $35.83{\scriptstyle\pm 0.73}$ & \cellcolor{blue!15}$61.82{\scriptstyle\pm 1.45}$ & \cellcolor{blue!15}$69.16{\scriptstyle\pm 1.90}$ & 
	   $81.35{\scriptstyle\pm 5.47}$ &
	   $76.03{\scriptstyle\pm 1.16}$ & $89.74{\scriptstyle\pm 0.55}$ & $87.28{\scriptstyle\pm 1.01}$ & 
	   \cellcolor{blue!15}74.32\\	   
	   {$Gain$}  & $\uparrow 43.17\%$ & $\uparrow 38.04\%$ & $\uparrow 18.41\%$ & $\uparrow 67.58\%$ & $\uparrow 15.61\%$ & $\uparrow 42.64\%$ 
	   & $\downarrow 0.85\%$ & $\uparrow 2.70\%$ & $\rightarrow 0.00\%$ & $\uparrow 20.61\%$\\
	   \bottomrule
    \end{tabular}
    \end{adjustbox}
    %\vspace{-0.5cm}
\end{table*}

\subsection{Experimental Setup}
\subsubsection{Datasets}
% \noindent
% \textbf{Datasets.} 
We evaluate the performance on nine well-known real-world node classification datasets, including three homophily datasets (Citeseer, Pubmed, and Cora) and six heterophily datasets (Texas, Wisconsin, Actor, Squirrel, Chameleon, and Cornell).
For all benchmarks, we use the same feature vectors, graph structure, class labels, and standard 10 fixed random splits (48\%/32\%/20\% of nodes per class for train/validation/test) provided in literature~\cite{pei2019geom,zhu2020beyond}.

\begin{itemize}
    \item {Homophily Datasets}
    \begin{itemize}
        \item \textit{Citeseer, Pubmed, Cora}~\cite{GCN}: For the basic citation datasets~\cite{sen2008collective}, nodes correspond to papers; edges correspond to citation links; the sparse bag-of-words are the feature representation of each node. Finally, the label of each node represents the topic of the paper. 
    \end{itemize}
    \item {Heterophily Datasets}
    \begin{itemize}
        \item \textit{Texas, Wisconsin, Cornell}~\cite{pei2019geom}: Cornell, Texas, and Wisconsin are the web page networks captured from the computer science departments of these universities in the WebKB dataset. In these networks, nodes and edges represent the web pages and hyperlinks. Similar to the Citations networks, words in the web page represent the node features in the bag-of-word form. The web pages are labeled into five categories: student, project, course, staff, and faculty.
        \item \textit{Squirrel, Chameleon}~\cite{pei2019geom}: Chameleon and Squirrel are web pages extracted from different topics in Wikipedia~\cite{rozemberczki2021multi}. Similar to WebKB, nodes and edges denote the web pages and hyperlinks among them, respectively, and informative nouns in the web pages are employed to construct the node features in the bag-of-word form. Webpages are labeled in terms of the average monthly traffic level.
        \item \textit{Actor}~\cite{pei2019geom}: The actor network contains the co-occurrences of actors in films, which are extracted from the heterogeneous information networks. It describes the complex relationships among films, directors, actors and writers~\cite{tang2009social}. In this network, nodes and edges represent actors and their co-occurrences in films, respectively. The actor’s Wikipedia page is exploited to extract features and node labels.
    \end{itemize}
\end{itemize}

\subsubsection{Baselines}
% \vspace{2pt}
% \noindent
%  \textbf{Baselines.} 
 We compare our method with the following baselines: \textbf{(1)} classical standard GNNs: GCN~\cite{GCN}, GAT~\cite{GAT} and GIN~\cite{xu2018powerful}; \textbf{(2)} recent state-of-the-art GNNs of specific structure tackling heterophily: Geom-GCN~\cite{pei2019geom}, MixHop~\cite{abu2019mixhop}, H2GCN~\cite{zhu2020beyond}, GPRGNN~\cite{chien2020adaptive}, FAGCN~\cite{bo2021beyond},
 GCN-Cheby~\cite{defferrard2016convolutional},
 JK-Net~\cite{xu2018representation}, 
 GCNII~\cite{chen2020simple};
 and \textbf{(3)} standard 2-layer MLP. To show the effectiveness and generalizability of our framework, we choose three simple GCN, GAT, and GIN as the graph convolution component in CAGNNs. For ease of comparison, we use the reported results of baselines in the literature~\cite{pei2019geom,zhu2020beyond}. Moreover, for the missing results under these splits, we rerun the released code 10 times and report the mean and standard deviation.
 
\begin{itemize}
    \item {Classical GNNs}
    \begin{itemize}
        \item \textit{GCN}~\cite{GCN}: GCN can be seen as a Laplacian smoother since it uses the mean aggregator to smooth each node and its neighbor's features.
        \item \textit{GIN}~\cite{xu2018powerful}: GIN utilizes the MLP to model the injective function when aggregation and generalizes the WL test.
        \item \textit{GAT}~\cite{GAT}: GAT is a graph neural network that applies the attention mechanism on node features to learn edge weights for aggregation.
    \end{itemize}
    \item {Heterophily GNNs}
    \begin{itemize}
        \item \textit{H2GCN}~\cite{bo2021beyond}: H2GCN proposed three designs with separate ego and neighbors, hider-order neighbors and a combination of intermediate representations to combine the message from neighbors.
        \item \textit{FAGCN}~\cite{bo2021beyond}: FAGCN divides the message from each edge into low-frequency and high-frequency signals during aggregation.
        \item \textit{GCN-Cheby}~\cite{defferrard2016convolutional}: GCN-Cheby combines higher-order neighbor information with Chebyshev polynomials from the spectral domain.
        \item \textit{GEOM-GCN}~\cite{pei2019geom}: GEOM-GCN utilizes structural similarity to capture the smooth structure in the latent space and long-range dependencies. 
        \item \textit{MixHop}~\cite{abu2019mixhop}: MixHop repeatedly mixes feature representations of neighbors at various distances to achieve higher-order message passing.
        \item \textit{GPRGNN}~\cite{chien2020adaptive}: GPRGNN modifies the convolution to the generalized page rank and learned an arbitrary polynomial graph filter to incorporate multi-scale information.
        \item \textit{JK-Net}~\cite{xu2018representation}: JK-Net combines intermediate node representations from each layer by concatenating them in the final layer.
        \item \textit{GCNII}~\cite{chen2020simple}: The state-of-the-art deep model combines initial connections and identity mapping to train a very deep GCN.
    \end{itemize}
\end{itemize}

\subsubsection{Hyper-parameters}
We adopt the same set of default hyper-parameters (2 layers and 64 hidden dimensions) for GCN, GAT, and GIN and the corresponding \ourModelName, which is the most widely used hyper-parameters setting as~\cite{bo2021beyond,chen2020simple,GAT}. For \ourModelName, we add only one linear layer with 128 hidden units. We employ the Adam optimizer and select the learning rate $\in \left \{ 0.001, 0.01, 0.05\right \}$, weight decay $\in \left \{ 0.00005, 0.0005 \right \}$ and dropout rate $\in \left \{ 0, 0.5 \right \}$ based on the validation sets.
For other models, we utilize their best default parameters in the original papers. 

\subsection{Performance Comparison with SOTA}\label{classification-comparison}
We report and compare the performance for the standard node classification task in Table~\ref{tab:ssnc-results}. For the classic GNNs (GCN, GAT and GIN), we first note that they outperform the MLP in the homophily datasets. It indicates that the homophily assumption and connected intra-class neighbors provide helpful node classification information. However, we also notice that they sometimes also perform better than the MLP under heterophily datasets. For instance, in the Actor dataset, the performance of MLP is approximately 36\%, but the traditional GCN only obtains 30\%. However, in the Chameleon dataset, the performance comparison between GCN and MLP is (60\% vs. 46\%). Hence, it implies that not all the inter-class edges in the heterophily datasets are harmful to the node classification.

In addition, we note that the GNNs with specific designs for the heterophily datasets outperform the traditional GNNs with a large margin for six heterophily datasets. Most of them (e.g., MixHop, JK-Net, and H2GCN) explicitly aggregate high-order neighbors' information to avoid the harmfulness of inter-class edges. For instance, the strong baseline H2GCN proves that the neighbor's high-order information is expectedly homophily-dominant and achieves 70.87\% average performance. In comparison, the average performance of the two-layer GCN is only 61.62\%.

In contrast, instead of considering that the inter-class edges are all harmful, we take each node's entire neighbor effect into account. As a result, compared with recently state-of-the-art heterophily GNNs, our \ourModelName can help traditional standard GNNs achieve competitive results while maintaining the performance on three homophily datasets.
Moreover, from the perspective of the spectral domain, CAGNN can help traditional GNNs adaptively learn different coefficients of K-order polynomial graph filters for each node. Therefore, our framework performs better than the spectral GNNs sharing the same polynomial filter coefficients for all nodes (e.g., GPRGNN and GCNII). Under our framework, the average performance on nine datasets of GIN, GAT, and GCN outperforms all the baselines, and the average performance gains are 9.81\%, 25.81\%, and 20.61\%, respectively.
Among them, the proposed \ourModelName with 2-layer GCN achieves the best average performance (74.32\%) over all datasets. 
It verifies the effectiveness of decoupling design and consideration of the neighbor effect when performing graph convolution.

\subsection{Relation between the Metrics and Performance}\label{metrics_vs_performance}

Table~\ref{tab:ssnc-results} also shows the different metrics for all datasets. 
All the metrics range from $[0,1]$ and a higher score of $\mathcal{H}_{\text {node }}$, $\mathcal{H}_{\text {edge}}$ and $\mathcal{H}_{\text {agg}}$ denotes higher homophily. 
However, a higher $\mathcal{H}_{\text {neighbor}}$ means lower identifiability of neighbors' distribution and a more challenging dataset, i.e., the neighbor distribution provides less useful information for classification. Neighbor distribution identifiability offers an alternative perspective to the inter-class edges approach ($\mathcal{H}_{\text {node }}$ and $\mathcal{H}_{\text {edge}}$) for understanding the heterophily problem in GNNs.
Note that only our $\mathcal{H}_{\text {neighbor}}$ can distinguish the dataset Actor (0.98) from others, in which the neighbors' distribution is nearly random and the best test classification accuracy is very low, i.e., 35.86. Our metric shows that the GNN does not outperform MLP (35.73) in the Actor datasets with randomness neighbor distribution. Therefore, it may be used as a guide to choosing whether to employ GNNs rather than MLP. 

\begin{table}[!t]
\centering
    \caption{Kendall correlation between different metrics and the performance of $\text{CAGNN}_{GCN}$. The higher coefficient and the lower p-value are better and more significant.}
    \label{tab:kendall}
    \setlength{\tabcolsep}{2.5mm}{
        \begin{tabular}{l|l|llll}
        \toprule
            Datasets & Kendall     & $\mathcal{H}_{\text {node }}$ & $\mathcal{H}_{\text {edge }}$ & {$\mathcal{H}_{\text {agg }}$} & $\mathcal{H}_{\text {neighbor}}$ \\
        \midrule
         \multirow{2}*{$>$500 nodes} &  coefficient & 0.733 & 0.828 & 0.467 & \textbf{0.867}\\ 
         ~ & p-value & 0.056 & 0.022 & 0.27 & \textbf{0.017}\\    
         \midrule 
         \multirow{2}*{All datasets} & coefficient & 0.11 & 0.25 & 0.44 & \textbf{0.59}\\   
         ~ & p-value & 0.7 & 0.34 & 0.12 & \textbf{0.02}\\    
        \bottomrule
    \end{tabular}}
\end{table}

Moreover, for the graph-level metrics ${\mathcal{H}_\text{neighbor}}$, we also report the Kendall rank correlation coefficient between these metrics and the performance of CAGNN in Table~\ref{tab:kendall}. The ${\mathcal{H}_\text{neighbor}}$ is more correlated with the performance over different datasets. Therefore, the proposed metric can be considered a rough evaluator to measure the difficulty of graphs for the node classification task using GNNs, especially for medium-scale graphs with more than 500 nodes.
We must have a sufficient number of nodes to have meaningful statistical information.

\begin{figure}[t]
    \centering
    \subfigure[Cora]{
    \begin{minipage}{0.45\linewidth}
      \centering
      \includegraphics[width=\linewidth]{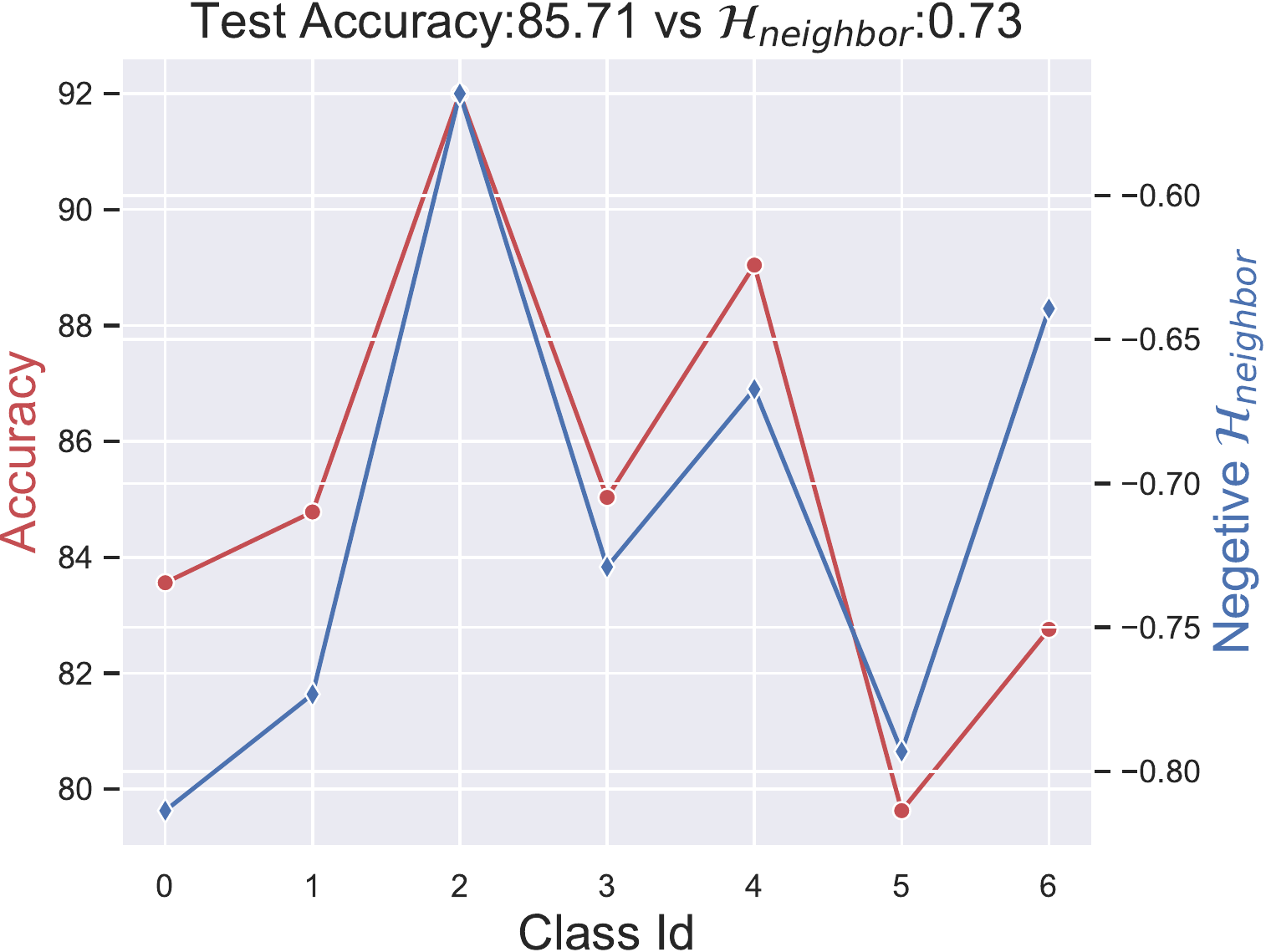}
      \label{fig:Cora-metric}
    \end{minipage}
    }
    \subfigure[Chameleon]{
    \begin{minipage}{0.45\linewidth}
      \centering
      \includegraphics[width=\linewidth]{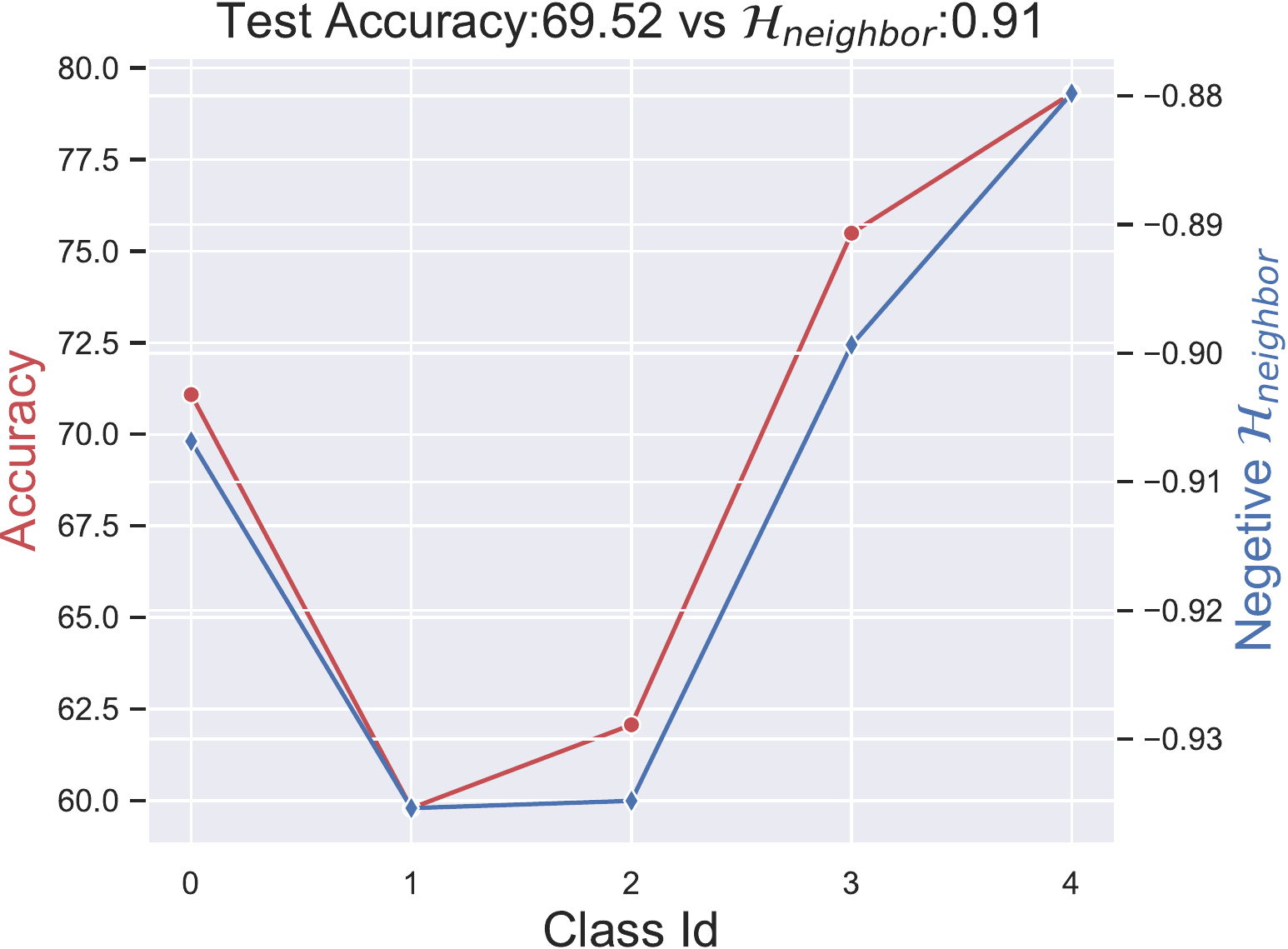}
      \label{fig:Chameleon-metric}
    \end{minipage}
    }
    \subfigure[Citeseer]{
    \begin{minipage}{0.45\linewidth}
      \centering
      \includegraphics[width=\linewidth]{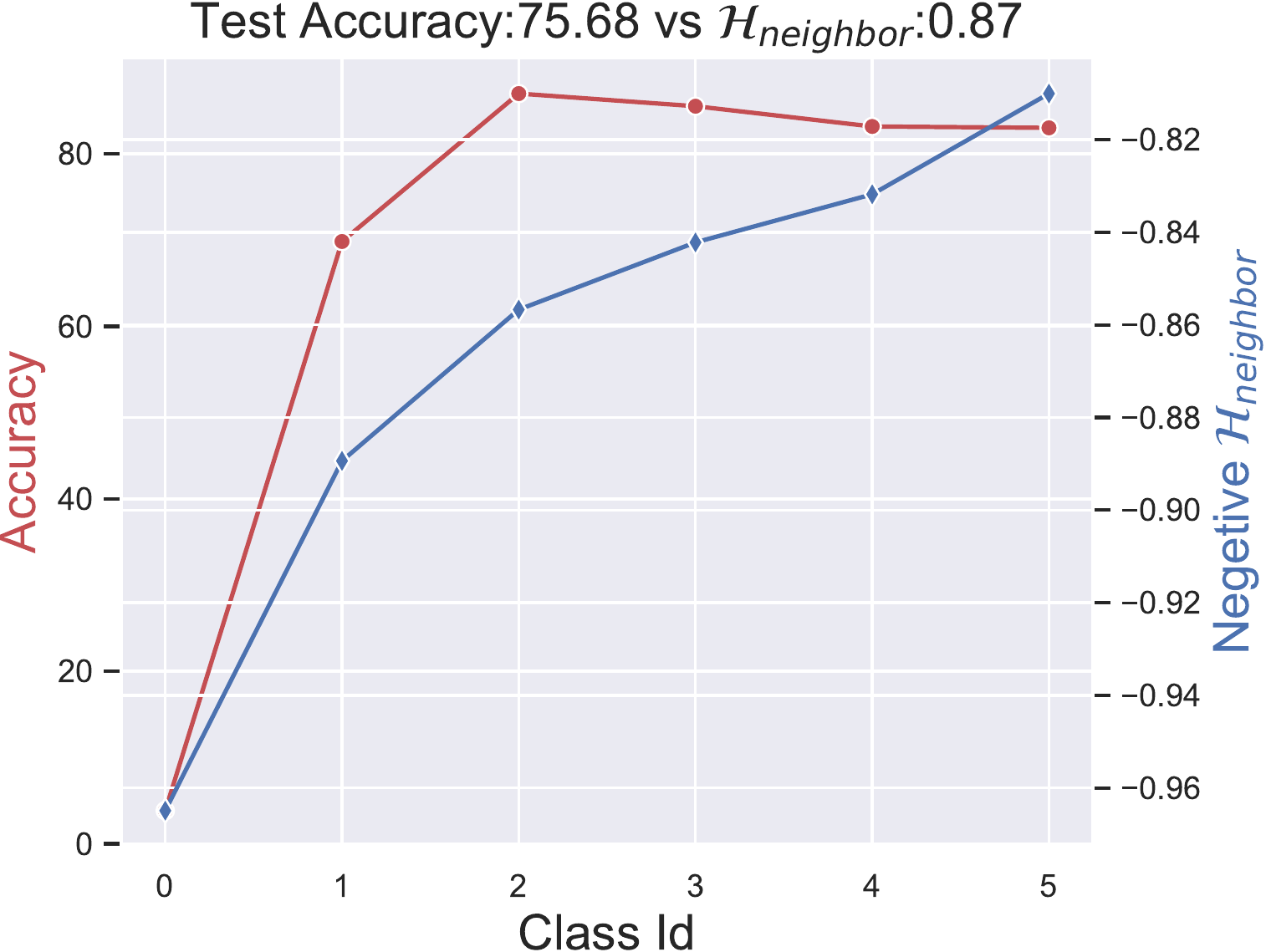}
      \label{fig:Citeseer-metric}
    \end{minipage}
    }
    \subfigure[Squirrel]{
    \begin{minipage}{0.45\linewidth}
      \centering
      \includegraphics[width=\linewidth]{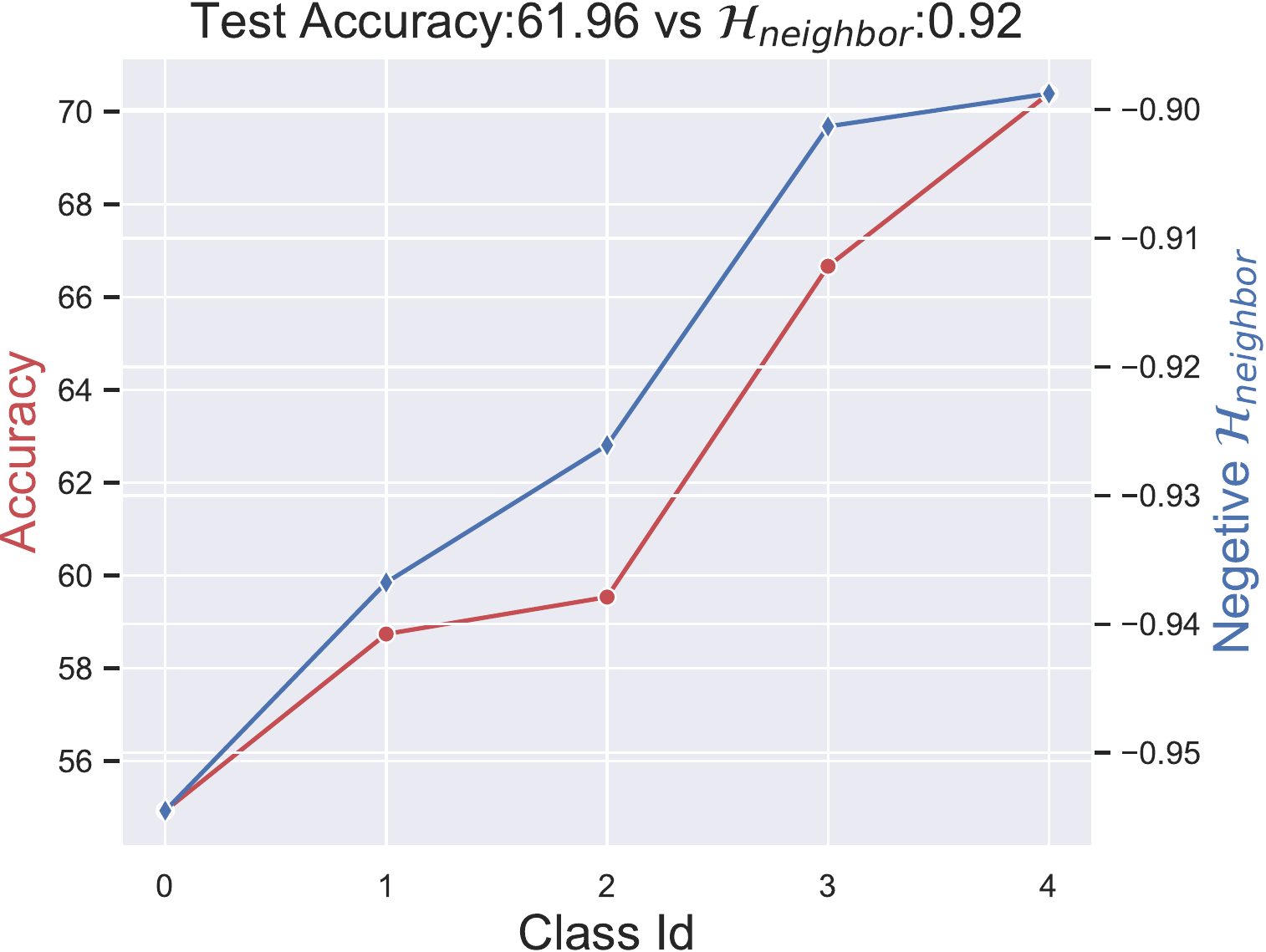}
      \label{fig:Squirrel-metric}
    \end{minipage}
    }
    \subfigure[Pubmed]{
    \begin{minipage}{0.45\linewidth}
      \centering
      \includegraphics[width=\linewidth]{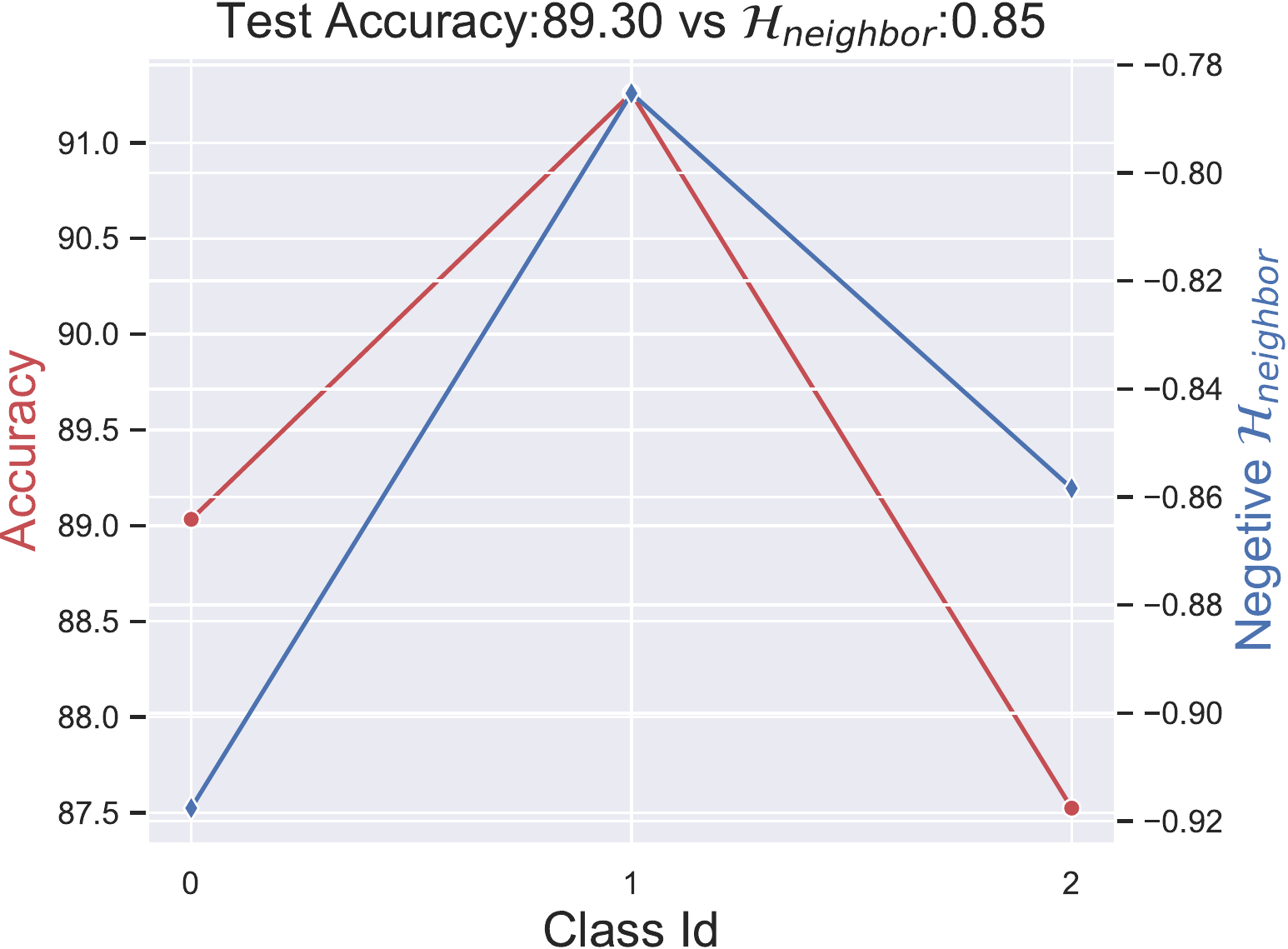}
      \label{fig:Pubmed-metric}
    \end{minipage}
    }
    \subfigure[Actor]{
    \begin{minipage}{0.45\linewidth}
      \centering
      \includegraphics[width=\linewidth]{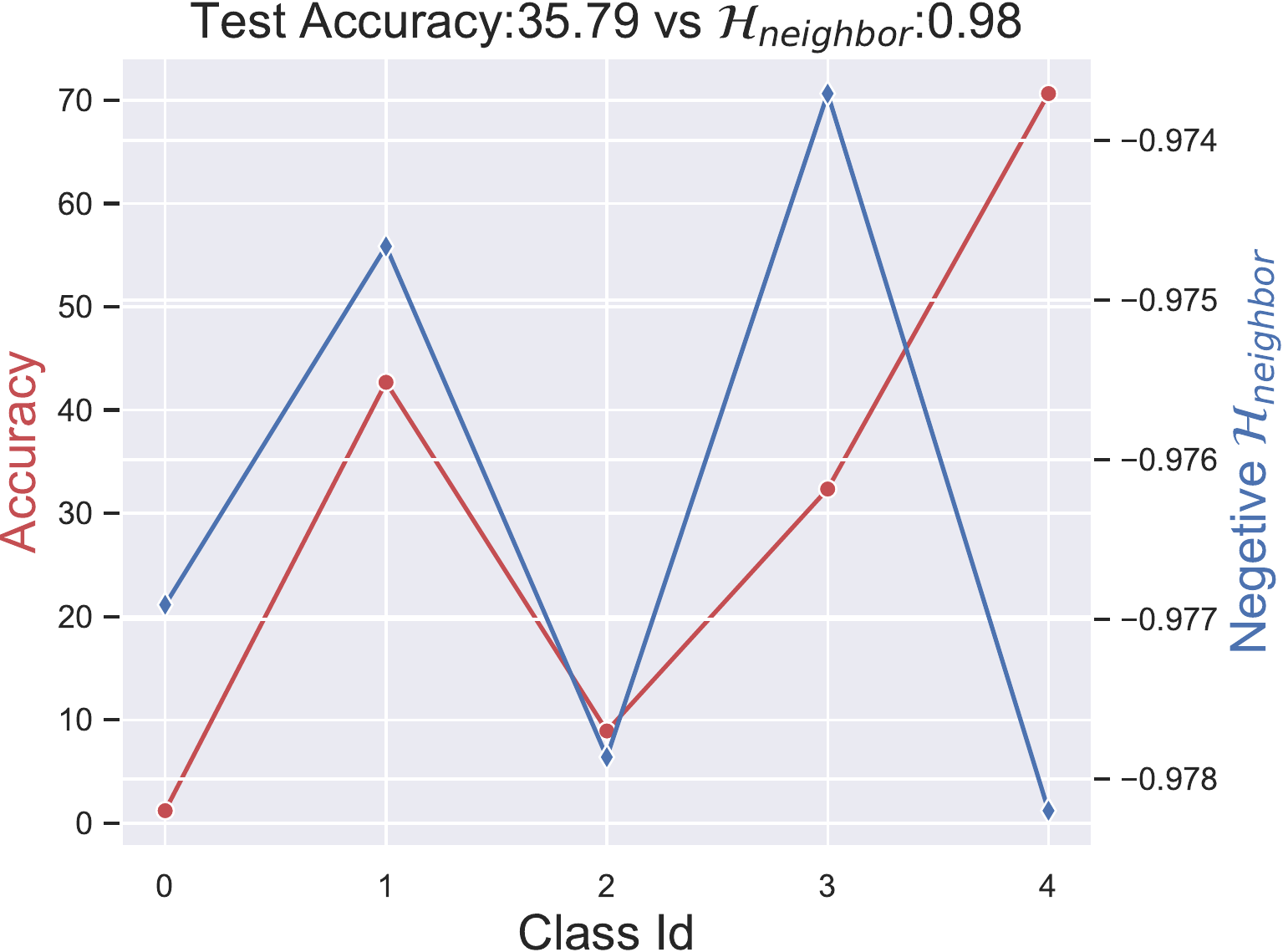}
      \label{fig:Actor-metric}
      
    \end{minipage}
    }
    \caption{Class-level comparison between our $\text{CAGNN}_{GCN}$ performance (red) and the negative class-level $\mathcal{H}_\text{neighbor}^c$ (blue) for Homophily (Cora, Citeseer, and Pubmed) and Heterophily (Chameleon, Squirrel, and Actor). The class-level performance of $\text{CAGNN}_{GCN}$ is highly consistent with the neighbor distribution identifiability for most datasets.}
    \label{fig:class-hete}
\end{figure}

For the class-level ${\mathcal{H}_\text{neighbor}^c}$, we show the relation with our CAGNN's performance under various datasets in Fig.~\ref{fig:class-hete}.  Since the larger ${\mathcal{H}_\text{neighbor}^c}$ means the lower identifiability, for clarity, we apply the negative ${\mathcal{H}_\text{neighbor}^c}$ for each class to show a positive correlation between the model's class-wise performance and the class-wise neighbor distribution identifiability. At the class level, we can see that the performance of $\text{CAGNN}_{GCN}$ is highly consistent with the neighbor distribution identifiability for most datasets, which indicates that the neighbor perspective can help explain the model performance on various datasets. It also verifies that our framework can adaptively evaluate the neighbor effect and guide each node to absorb helpful information for the downstream node classification task. 

\subsection{Ablation Study}
In this section, we compare CAGNN with its variants for Mixer and Normalization to validate the effectiveness of each component.
When testing different Mixer variants, we fix the Normalization to L2. Also, we set the Mixer to linear to test different Normalization variants. 
We select the best hyper-parameters of each variant and run experiments for each dataset under 10 random splits to report the average performance on all datasets.

\vspace{4pt}
\noindent
\textbf{Variants of Mixer.} The mixer module is a critical part of our framework to evaluate the \Q{neighbor effect} and feature fusion. We compared our results with six variants.
\begin{itemize}
    \item \textit{Add}: The Add mixer is implemented by ${\mathbf{S}}^{l} =\operatorname{Norm}( {\mathbf{S}}^{l-1}+{\mathbf{H}}^{l})$.
    \item \textit{Concat}: The Concat mixer is implemented by ${\mathbf{S}}^{l} = \operatorname{Norm}([{\mathbf{S}}^{l-1}||{\mathbf{H}}^{l}])$. Note that we only apply the Norm at the first and last layers for the Concat variant to maintain stability.
    \item \textit{Global}: The Global mixer is ${\mathbf{S}}^{l} = \operatorname{Norm}((1-\alpha^l){\mathbf{S}}^{l-1}+\alpha^l{\mathbf{H}}^{l})$, where the learnable scalar $\alpha^l$ is shared for all nodes.
    \item {\textit{Unshared}}: {It replaces the shared-parameterized $\text{f}_{\operatorname{mixer}}$ to unshared version for each layer.}
    \item \textit{MLP-2/3}: It replaces the linear layer in $\text{f}_{\operatorname{mixer}}$ with two or three layers of shared-parameterized MLP.
\end{itemize}

\begin{table}[t]
    \centering
    \small
    \caption{The average performance (test accuracy) over all datasets for the ablation study of different types of Mixers and Normalization.}
    \label{tab:ablation}
\begin{tabular}{l|c|ccc}
\toprule
 & Variants & $\text{CAGNN}_{GIN}$ & $\text{CAGNN}_{GAT}$ & $\text{CAGNN}_{GCN}$ \\ \midrule

\multirow{3}{*}{Mixer} & Add & 71.91 & 70.63 & 71.41 \\ 
 & Concat & 69.50 & 71.27 & 72.23 \\ 
 & Global & 72.17 & 72.52 & 73.43 \\ 
 & {Unshared} & 72.82 & 73.06 & 73.88\\ 
 & MLP-2 & 72.84 & 73.15 & \bf{74.38} \\  
 & MLP-3 & 72.57 & 73.47 & 74.20 \\  \midrule

\multirow{3}{*}{Norm} 
& None & 72.29 & 73.03 &  73.01\\
 & BatchNorm & 65.11 & 61.61 & 64.92  \\  
 & LayerNorm &  72.41 & 72.19 &  72.45 \\ \midrule

  & Ours & \textbf{73.28} & \textbf{73.67} & 74.32 \\ \bottomrule
\end{tabular}
\end{table}

From the results of these mixer variants in Table~\ref{tab:ablation}, we have the following observations: (1) All these variants of mixer can substantially improve the performance of standard GNNs, which verifies the effectiveness of the decoupled design. (2) The results of the global mixer are better than those of the Add and Concat mixer, indicating that the neighbor effect is diverse under different datasets. (3) Our 1-layer mixer and multilayer MLP variants achieve consistently better performance than the global model. These mixers can adaptively learn each node's neighbor effect, which verifies that the learnable node-level neighbor effect plays a more vital role than the global mixer.
(4) Compared with the {unshared-parameterized version} and the multiple layer MLPs in the current $\text{f}_{\operatorname{mixer}}$ module, the result shows that the current shared-parameterized one-layer mixer is simple yet effective.

\vspace{4pt}
\noindent
\textbf{Variants of Normalization.} We also compare the L2 norm for each node with three variants (None, BatchNorm~\cite{Ioffe2015BatchNA} and LayerNorm~\cite{Ba2016LayerN}) to show its effectiveness. 
We first observe that the None normalization version achieves comparable performance, demonstrating that our framework is stable and competitive. 
However, the BatchNorm usually assumes the independent and identical distribution of each sample in deep learning, which may not be reasonable for the heterophilic graph and results in worse performance. Moreover, LayerNorm is also not beneficial since the layer norm introduces learning parameters that may be redundant for GNNs. Compared with these variants, our simple yet effective L2 Norm achieves the best average performance.

\subsection{Robustness Analysis}
To investigate whether the proposed CAGNN framework can help basic GNNs become more robust, we report the performance comparison between the basic GNNs and corresponding CAGNN in the over-smoothing and noisy edges scenario.

\begin{figure}[t]
\subfigure[Robustness of oversmooth]{
\begin{minipage}[t]{0.45\linewidth}
\centering
\includegraphics[width=\linewidth]{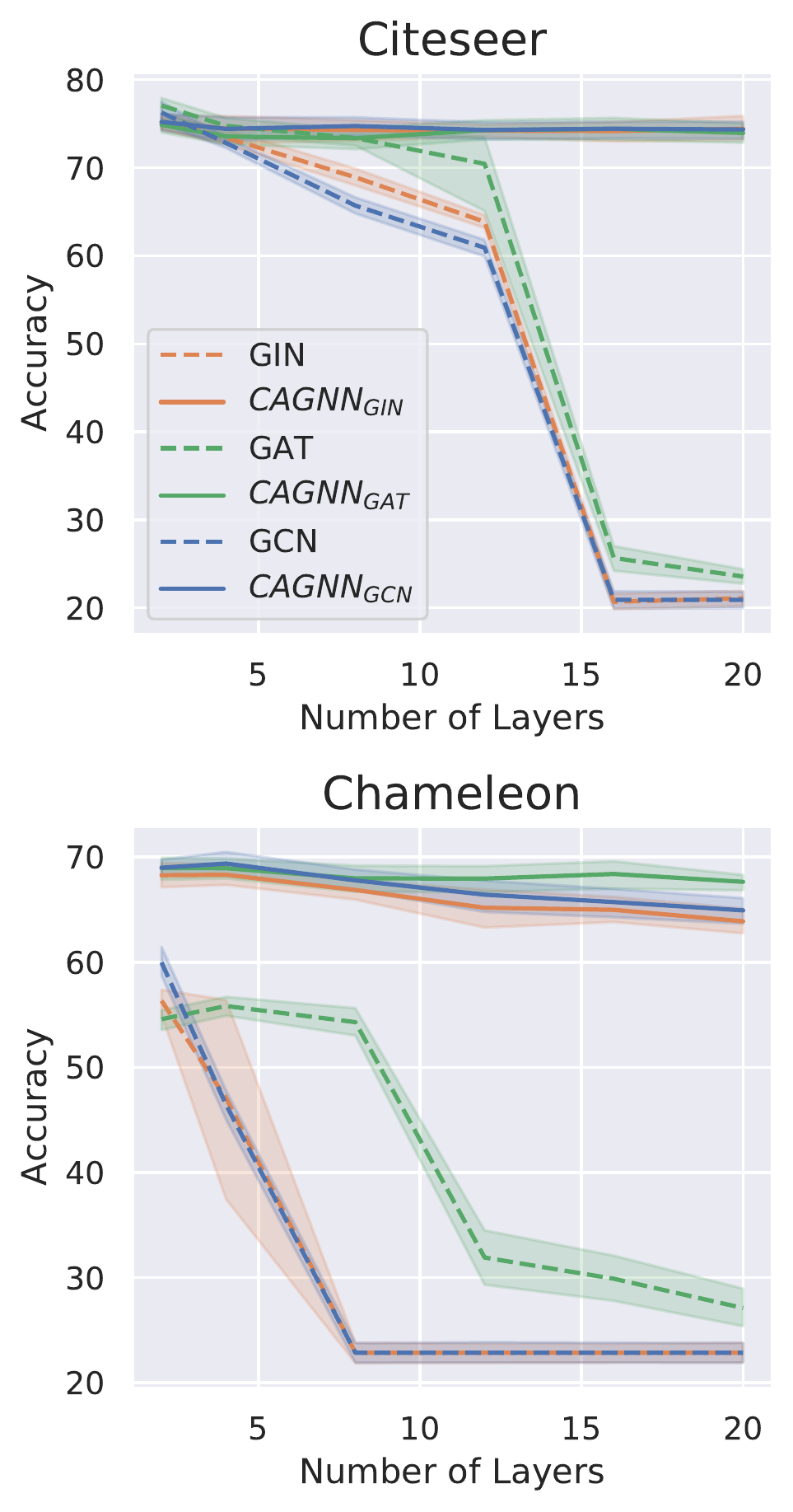}
\label{fig:oversmooth}
\end{minipage}
}
\subfigure[Robustness of noisy edges]{
\begin{minipage}[t]{0.45\linewidth}
\centering
\includegraphics[width=\linewidth]{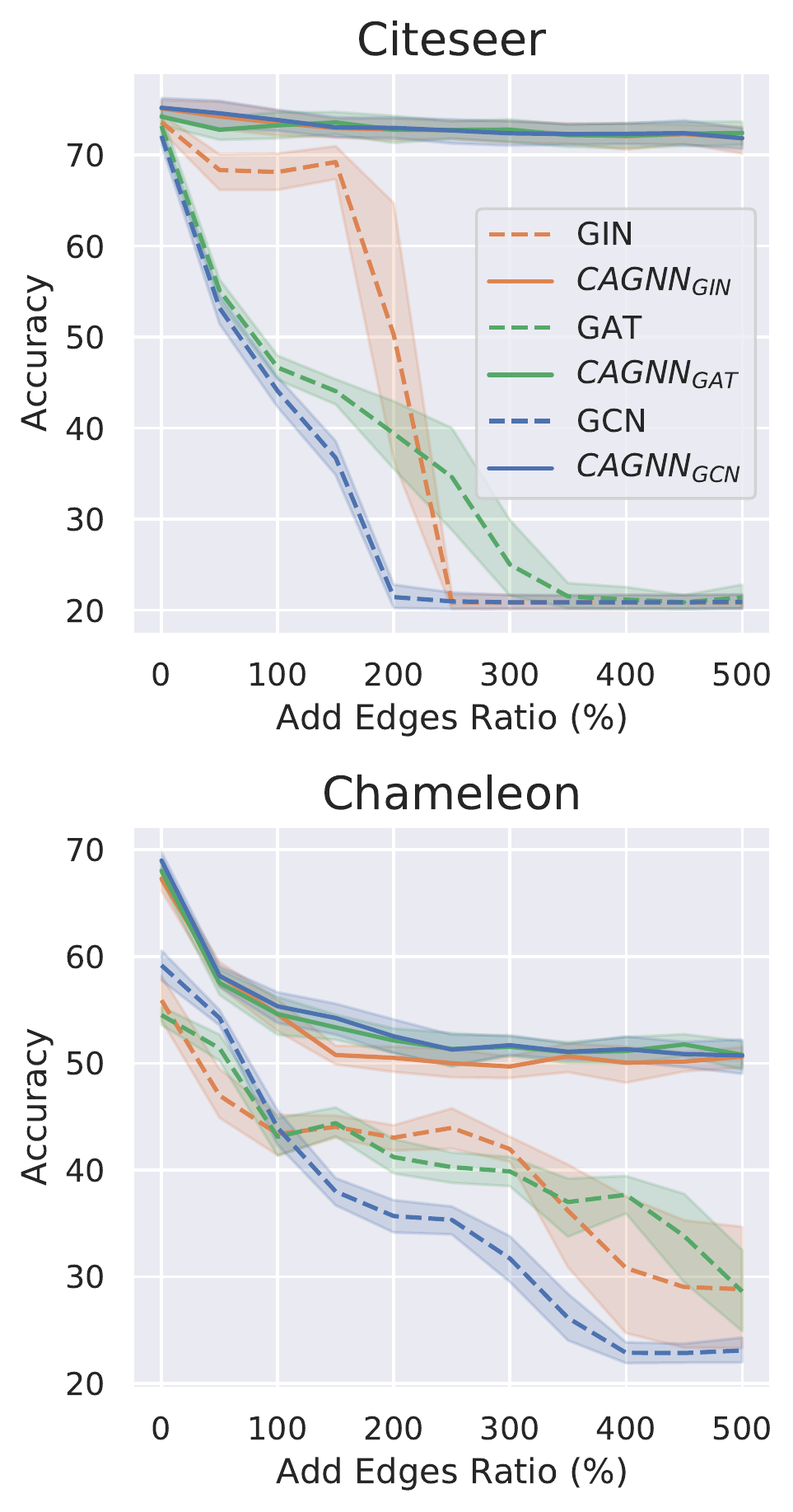}
\label{fig:noisy-graph}
\end{minipage}
}
\caption{Performance comparison with traditional GNNs and the variants under the proposed \ourModelName framework in terms of (a) {\it number of layers}, and (b) {\it ratio of adding noisy edges} on Homophily (Citeseer) and Heterophily (Chameleon) datasets.}
\label{fig:robutness}
\end{figure}

\vspace{4pt}
\noindent
\textbf{Alleviating oversmooth.} It is well known that the traditional graph convolution is sensitive to the number of convolution layers due to the over-smoothing problem~\cite{li2018deeper}. Our \ourModelName are able to increase the robustness of traditional graph convolution to avoid over-smoothing. 
As shown in Fig.~\ref{fig:oversmooth}, when the number of layers increases, the performance of traditional GNNs drops rapidly due to over-smoothing.
Moreover, due to the massive inter-class edges, the over-smoothing phenomenon occurs earlier in the heterophily datasets.
In contrast, the methods under the proposed framework are more stable and more consistent on both the Homophily (Citeseer) and Heterophily (Chameleon) datasets. 
The reason is that our framework has the ability to avoid incorporating the over-smoothing features to maintain the discrimination power for each node.

\vspace{4pt}
\noindent
\textbf{Alleviating noisy edges.} Most GNNs are also sensitive to the noisy edges in graphs~\cite{franceschi2019learning,xutopology}. To evaluate the robustness of the proposed framework on noisy graphs, we construct graphs with random edge addition according to the literature~\cite{chen2020iterative}. Specifically, we randomly add 
25\%$\sim$500\%
edges in the original graphs. As shown in Fig.~\ref{fig:noisy-graph}, our \ourModelName achieve significantly better prediction for noisy graphs compared with the basic GNNs. It also demonstrates that our decouple design and the mixer module are able to learn to discard the noisy features from neighbors.

\subsection{Visualization and Interpretability}
\begin{figure}[t]
  \centering
  \centerline{
  \includegraphics[width=\linewidth]{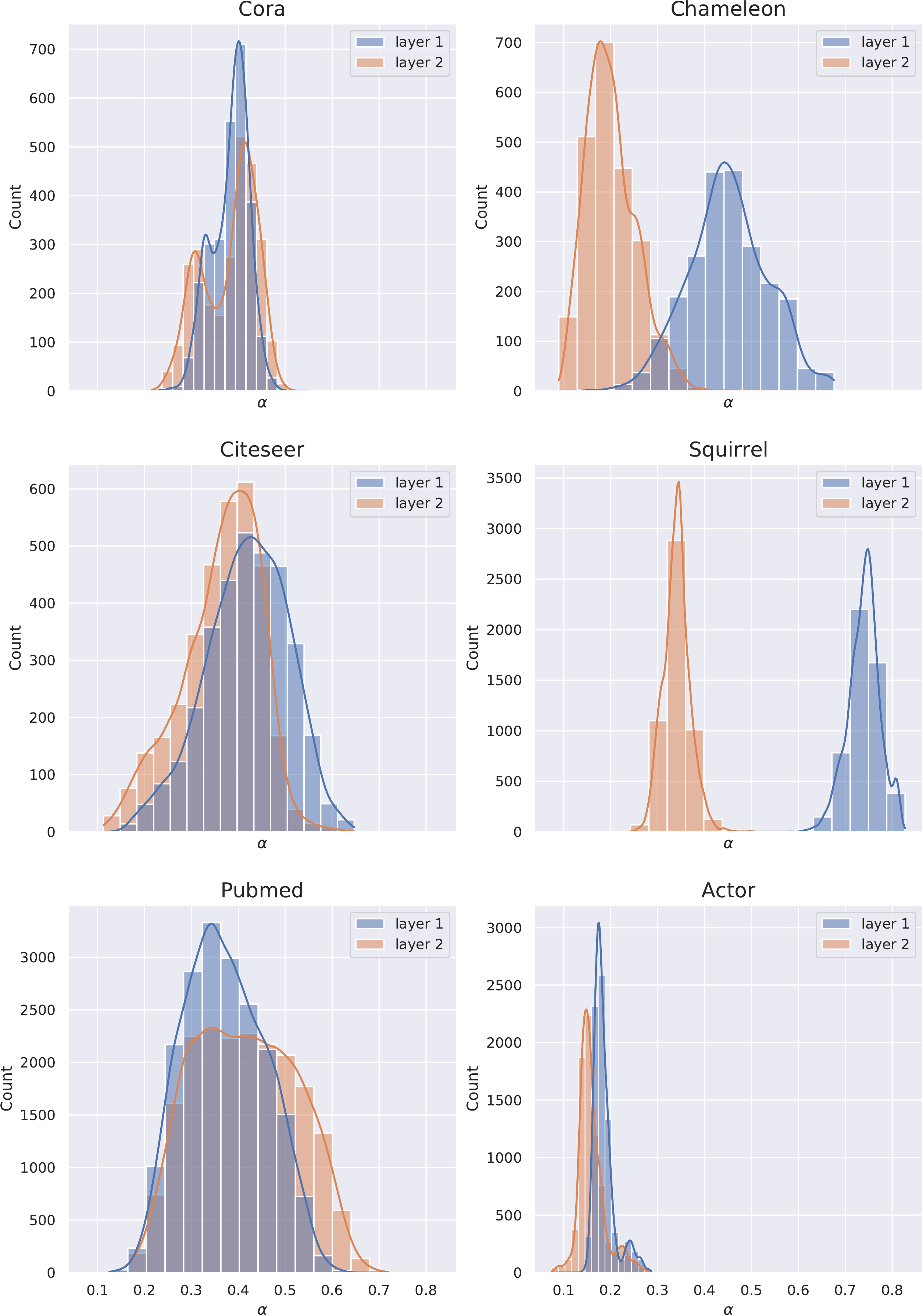}
}
\caption{Visualization of nodes' neighbor importance scores of the first and second layers on different datasets.}
\label{fig:visual-gate}
\end{figure}

To verify whether \ourModelName can adaptively learn the different \Q{neighbor effect} of each node, we visualize the neighbor importance score $\bm{\alpha}$ distribution of each layers on both homophily and heterophily datasets, where the results are shown in Fig.~\ref{fig:visual-gate}.
We have the following observation:
(1) As we can see, for homophily datasets (Cora, Citeseer, and Pubmed), the coefficients are near 0.4 for most nodes. This indicates that the information from neighbors is helpful for the downstream classification, which is consistent with the homophily assumption. 
(2) However, we observe similar trends in heterophily datasets (Chameleon and Squirrel) that most nodes still absorb the neighbors' information. Moreover, the distribution of the second layer score shows that 1-hop neighbors are more important than 2-hop neighbors. This phenomenon is contrary to the previous studies that the inter-class edges are all harmful, which implies that graph convolution can still extract classification information from an inter-class neighbor with a non-random distribution.
(3) The Actor dataset has a similar proportion of inter-class edges with Chameleon and Squirrel, but the neighbor distribution is nearly random ($\mathcal{H}_\text{neighbor}$ is 0.98). 
The neighbor importance score of nodes tends to be 0, which verifies that the inter-class edges with random distribution are harmful and guides the model to discard neighbors' information when aggregation. 

\subsection{Efficiency}
\begin{figure}[t]
  \centering
  \centerline{
  \includegraphics[width=8cm]{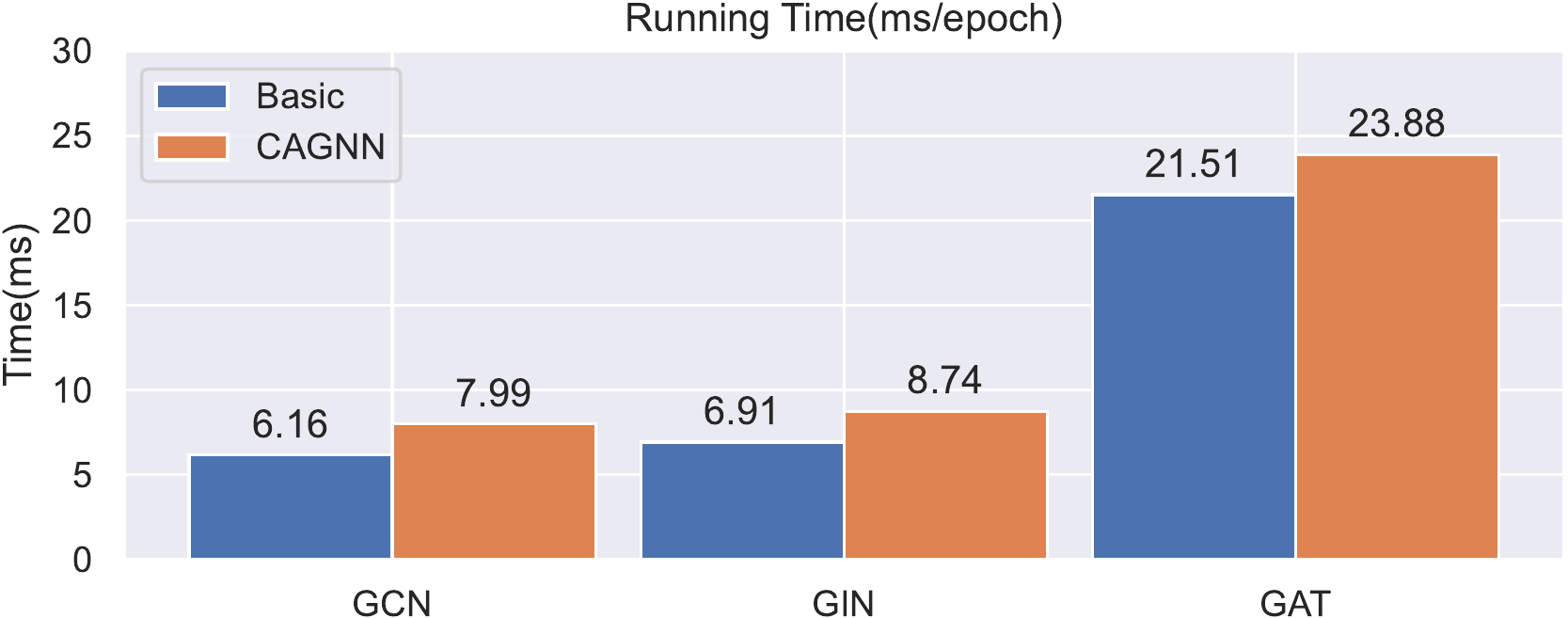}}
\caption{Average running time per epoch (ms)}
\label{fig:efficiency}
\end{figure}

Finally, we investigate the complexity of the proposed framework. The complexity of computing the neighbor importance score and mixing for all nodes is $O(Nd)$ where $N$ 
is the total number of nodes, and $d$ is the dimension of the hidden nodes. The computational complexity is on par with the neighborhood aggregation operation in GNNs, which is also $O(Nd)$. Hence, the complexity of our method and classic GNNs are on the same level.
We also report the average training time of each epoch over all datasets of the standard graph convolution and our method in
Fig.~\ref{fig:efficiency}. It indicates that our model scales similarly to the basic GNNs with a small computation cost for the mixer module.

\section{{Conclusion}}
In this paper, we have investigated the neighbor effect on heterophilic datasets.
Unlike previous works that argue that inter-class edges are harmful for node classification, we find that inter-class edges can be helpful when the neighbor distribution is identifiable. Furthermore, instead of computing node/edge-level metrics for a graph with heterophily, we propose a new measurement from the entire neighbor-level perspective via von Neumann entropy.
The proposed new metric sheds new light on the heterophily problem. It enables us to explain the performance variation of GNNs for different datasets and can be used to guide the application of GNNs.

We also proposed a simple yet effective heterophily GNN framework, CAGNN, which adds just one mixer layer to enhance the performance of conventional GNNs. The node features are first decoupled into discrimination and aggregation parts, and then adaptively fused according to the neighbor effect of each node. Our experiments on nine well-known benchmark datasets not only demonstrate the effectiveness of the proposed framework as a plug-in to consistently improve existing GNNs on heterophilic graphs, but also indicate good interpretability for determining whether neighbor information is helpful for the downstream node classification task.

Similar to other node- and edge-level metrics, our metrics are constrained by the requirement for node label information. Therefore, it is non-trivial to explicitly combine these metrics to guide the aggregation process during training without label information. One of our future works is to incorporate the pseudo-label with these metrics to improve the aggregation of GNNs. Besides, beyond node classification, exploiting the connection between neighbor identifiability with the edge or graph-level task is also an interesting topic for future work.
%\input{sections/Appendix}
% \vfill\pagebreak
\IEEEpeerreviewmaketitle

\bibliographystyle{IEEEtran}
\bibliography{ref}

\vfill
\end{document}